\documentclass[letterpaper]{article} 
\usepackage{aaai22}  
\usepackage{times}  
\usepackage{helvet}  
\usepackage{courier}  
\usepackage[hyphens]{url}  
\usepackage{graphicx} 
\usepackage{natbib}  
\usepackage{caption} 
\DeclareCaptionStyle{ruled}{labelfont=normalfont,labelsep=colon,strut=off} 
\usepackage{tabularx}
\usepackage{xparse}
\usepackage{amsmath}
\usepackage{amsthm}
\usepackage{amssymb}
\usepackage{xspace}
\usepackage{relsize}
\usepackage{multirow}
\usepackage{comment}
\usepackage{adjustbox}
\usepackage{math_commands}
\usepackage{manfnt}
\usepackage{scalerel}
\usepackage{algorithm}
\usepackage[noend]{algorithmic} 

\usepackage{booktabs}   

\usepackage{stackengine}
\stackMath
\newcommand\tsup[2][2]{%
 \def\useanchorwidth{T}%
  \ifnum#1>1%
    \stackon[-.5pt]{\tsup[\numexpr#1-1\relax]{#2}}{\scriptscriptstyle\sim}%
  \else%
    \stackon[.5pt]{#2}{\scriptscriptstyle\sim}%
  \fi%
}

\renewcommand{\tilde}[1]{\tsup[1]{#1}}
\newcommand{\uline}[1]{\underline{#1}}
\newcommand{\pddl}[1]{\textsf{\small #1}}

\newcommand{\function}[1]{\textsc{#1}}

\newcommand{\mycolor}[2]{\textcolor{#1}{#2}}

\newcommand{\red}[1]{\mycolor{red}{#1}}
\newcommand{\green}[1]{\mycolor{ForestGreen}{#1}}

\newcommand{\spc}[2][c]{%
  \begin{tabular}[#1]{@{}c@{}}#2\end{tabular}}
\def\_{\\[-0.3em]}
\def\spm{$\mathsmaller{\mathsmaller{\pm}}$}

\makeatletter
\let\@myref\ref

\newcommand{\refsec}[1]{Section \@myref{#1}}
\newcommand{\refseq}[1]{Sec.\,\@myref{#1}}
\newcommand{\refsecs}[2]{Sec. \@myref{#1}-\@myref{#2}}
\newcommand{\refig}[1]{Figure \@myref{#1}}
\newcommand{\refigs}[2]{Figure \@myref{#1}-\@myref{#2}}
\newcommand{\reftbl}[1]{Table \@myref{#1}}
\newcommand{\refstep}[1]{Step \@myref{#1}}
\newcommand{\refalgo}[1]{Algorithm \@myref{#1}}
\newcommand{\refchap}[1]{Chap.\,\@myref{#1}}
\newcommand{\reflst}[1]{List \@myref{#1}}
\newcommand{\refeq}[1]{Eq.\,\@myref{#1}}
\renewcommand{\eqref}[1]{Eq.\,\@myref{#1}}

\newcommand{\refthm}[1]{Thm.\,\@myref{#1}}
\newcommand{\refline}[1]{line\,\@myref{#1}}
\newcommand{\reflines}[2]{lines\,\@myref{#1}-\@myref{#2}}

\newcommand{\refappx}[1]{Appendix \@myref{#1}}

\makeatother

\newcounter{list}[section]

\usepackage{natbib}

\usepackage[marginpar]{todo}          

\newlength{\maxwidth}
\newcommand{\algalign}[2]
{\makebox[\maxwidth][r]{$#1{}$}${}#2$}

\newcommand{\nlm}{\function{\small NLM}}

\newcommand{\pre}{\function{pre}}
\newcommand{\adde}{\function{add}}
\newcommand{\dele}{\function{del}}

\newcommand{\cost}{\function{cost}}

\def\hash{\text{\relsize{-1}\#}}
\newcommand{\ar}[1]{\hash{}#1}

\newcommand{\domind}{domain-independent\xspace}
\newcommand{\astar}{\xspace {$A^*$}\xspace}

\newcommand{\gbfs}{\function{gbfs}}

\newcommand{\qvalue}{action-value\xspace}
\newcommand{\qvalues}{action-values\xspace}

\makeatletter

\newcommand{\newheuristic}[2]{%
 \def#1{%
  \ifmmode%
  h^\text{#2}\xspace%
  \else%
  \text{#2}\xspace%
  \fi%
 }%
}

\newheuristic{\lmcut}{LMcut}
\newheuristic{\mands}{M\&S}
\newheuristic{\pdb}{PDB}
\newheuristic{\ff}{FF}
\newheuristic{\ce}{CEA}
\newheuristic{\cg}{CG}
\newheuristic{\ad}{add}
\newheuristic{\hmax}{max}
\newheuristic{\lc}{LC}
\newheuristic{\blind}{blind}

\newcommand{\newlearnedheuristic}[2]{%
 \def#1{%
  \ifmmode%
  H^\text{#2}\xspace%
  \else%
  \text{#2}\xspace%
  \fi%
 }%
}

\newlearnedheuristic{\Hlmcut}{LMcut}
\newlearnedheuristic{\Hmands}{M\&S}
\newlearnedheuristic{\Hpdb}{PDB}
\newlearnedheuristic{\Hff}{FF}
\newlearnedheuristic{\Hce}{CEA}
\newlearnedheuristic{\Hcg}{CG}
\newlearnedheuristic{\Had}{add}
\newlearnedheuristic{\Hmax}{max}
\newlearnedheuristic{\Hlc}{LC}
\newlearnedheuristic{\Hblind}{blind}

\newcommand{\newUnitCostHeuristic}[2]{%
 \def#1{%
  \ifmmode%
  \hat{h}^\text{#2}\xspace%
  \else%
  \text{#2}\xspace%
  \fi%
 }%
}

\newUnitCostHeuristic{\lmcuto}{LMcut}
\newUnitCostHeuristic{\mandso}{M\&S}
\newUnitCostHeuristic{\ffo}{FF}
\newUnitCostHeuristic{\ceo}{CEA}
\newUnitCostHeuristic{\cgo}{CG}
\newUnitCostHeuristic{\ado}{add}
\newUnitCostHeuristic{\gco}{GoalCount}
\newUnitCostHeuristic{\lco}{LC}

\makeatother

\renewcommand{\ref}[1]{\textbf{Do not use ``ref'' directly!}}

\usepackage[svgnames]{xcolor} 

\author{
Clement Gehring*\textsuperscript{\rm 1},
Masataro Asai*\textsuperscript{\rm 2},
Rohan Chitnis\textsuperscript{\rm 1},
Tom Silver\textsuperscript{\rm 1},\\
Leslie Pack Kaelbling\textsuperscript{\rm 1},
Shirin Sohrabi\textsuperscript{\rm 3},
Michael Katz\textsuperscript{\rm 3}
\\
}
\affiliations {
*: equal contributions.
\textsuperscript{\rm 1}MIT,
\textsuperscript{\rm 2}MIT-IBM Watson AI Lab,
\textsuperscript{\rm 3}IBM Research \\
\{gehring,lpk\}@csail.mit.edu,
\{ronuchit,tslvr\}@mit.edu,
\{masataro.asai,michael.katz1\}@ibm.com,
ssohrab@us.ibm.com
}

\title{Reinforcement Learning for Classical Planning:\\
Viewing Heuristics as Dense Reward Generators
}

\begin{document}
\maketitle
\begin{abstract}
  Recent advances in reinforcement learning (RL) have led to a growing
  interest in applying RL to classical planning domains or
  applying classical planning methods to some complex RL domains.
  However, the
  long-horizon goal-based problems found in classical planning lead to
  sparse rewards for RL, making direct application inefficient.
  In this paper, we propose to leverage
  domain-independent heuristic functions commonly used in the
  classical planning literature to improve the sample efficiency of RL.
  These classical heuristics act
  as dense reward generators to alleviate the sparse-rewards issue and
  enable our RL agent to learn domain-specific value functions as residuals on these heuristics, making learning easier.
  Correct application of this technique requires consolidating the discounted metric used in RL and the non-discounted metric used in heuristics.
  We implement the value functions
  using Neural Logic Machines, a neural network architecture designed for
  grounded first-order logic inputs. We demonstrate on several
  classical planning domains that using classical heuristics for RL allows for
  good sample efficiency compared to sparse-reward RL. We further show that our learned value functions
  generalize to novel problem instances in the same domain.
\end{abstract}

\section{Introduction}
\label{sec:introduction}

Deep reinforcement learning (RL) approaches have several strengths
over conventional approaches to decision making problems,
including compatibility with complex and unstructured observations,
little dependency on hand-crafted models, and some robustness to stochastic environments. However, they are notorious for their poor
sample complexity; e.g., it may require $10^{10}$ environment
interactions to successfully learn a policy for Montezuma's Revenge
\citep{badia2020agent57}. This sample inefficiency prevents their
applications in environments where such an exhaustive set of interactions is physically or
financially infeasible. The issue is amplified in
domains with sparse rewards and long horizons, where the reward signals for success are
difficult to obtain through random interactions with the
environment.

In contrast, research in AI Planning and classical planning has been primarily
driven
by the identification of tractable fragments of originally
PSPACE-complete planning problems
\citep{bylander1994},
and the use of the cost of the tractable \emph{relaxed} problem as
\emph{\domind} heuristic guidance
for searching through the state space of the original problem.
Contrary to RL approaches, classical planning has focused on
long-horizon problems with solutions well over 1000 steps
long \citep{jonsson2007role,Asai2015}.
Moreover, classical planning
problems inherently have sparse rewards ---
the objective of classical planning is to produce a sequence of actions that achieves a goal.
However,
although domain-independence is a welcome advantage,
domain-independent methods can be vastly outperformed by carefully engineered
domain-specific methods
such as a specialized solver for Sokoban \citep{junghanns00sokoban}
due to the no-free-lunch theorem for search problems \citep{wolpert1995no}.
Developing such domain-specific heuristics can require intensive
engineering effort, with payoff only in that single domain.
We are thus interested in developing domain-independent methods for
\emph{learning} domain-specific heuristics.

In this paper, we draw on the strengths of reinforcement learning and
classical planning to propose an RL framework for learning to solve
STRIPS planning problems. We propose to leverage classical heuristics,
derivable automatically from the STRIPS model, to accelerate RL agents
to learn a domain-specific neural network value function.
The value function, in turn, improves over existing heuristics and
accelerates search algorithms at evaluation time.

To operationalize this idea, we use \emph{potential-based reward shaping} \citep{ng1999policy},
a well-known RL technique with guaranteed theoretical properties. A key
insight in our approach is to see classical heuristic functions as
providing \emph{dense rewards} that greatly accelerate the learning
process in three ways. First, they allow for efficient, informative
exploration by initializing a good baseline reactive agent that
quickly reaches a goal in each episode during training. Second, instead of learning the value function directly, we learn a \emph{residual} on the heuristic value, making learning easier. Third, the learning agent
receives a reward by reducing the estimated cost-to-go (heuristic value). This effectively
mitigates the issue of sparse rewards by allowing the agent to receive
positive rewards more frequently.

We implement our neural network value functions as
Neural Logic Machines \cite[NLM]{dong2018nlm},
a recently proposed neural network architecture that can directly process
first-order logic (FOL) inputs, as are used in classical planning problems.
NLM takes a dataset expressed in grounded FOL representations and
learns a set of (continuous relaxations of) lifted Horn rules.
The main advantage of NLMs is that they structurally \emph{generalize} across different
numbers of terms, corresponding to objects in a STRIPS encoding.  Therefore, we find that our learned value functions are able to generalize effectively to problem instances of arbitrary sizes in the same domain.

We provide experimental results that validate the effectiveness of the proposed approach in 8 domains from
past IPC (International Planning Competition) benchmarks, providing detailed considerations on the reproducibility of the experiments.
We find that our reward shaping approach achieves good sample efficiency compared to sparse-reward RL,
and that the use of NLMs allows for generalization to novel problem instances.
For example, our system learns from \pddl{blocksworld} instances with 2-6 objects,
and the result enhances the performance of solving instances with up to 50 objects.

\section{Background}
\label{sec:background}

We denote a multi-dimensional array in bold.
$\va;\vb$ denotes a concatenation of tensors $\va$ and $\vb$ in the last axis.
Functions (e.g., $\log,\exp$) are applied to arrays element-wise.

\subsection{Classical Planning}
\label{sec:planning}

We consider planning problems in the STRIPS subset of PDDL
\cite{fikes1972strips}, which for simplicity we refer to as lifted
STRIPS. We denote such a planning problem as a 5-tuple
$\brackets{O,P,A,I,G}$.  $O$ is a set of objects, $P$ is a set of
predicates, and $A$ is a set of actions.
We denote the arity of predicates $p\in P$ and action $a\in A$ as $\ar{p}$ and $\ar{a}$, and their parameters as, e.g., $X=(x_1,\cdots,x_{\ar{a}})$.
We denote the set of predicates and actions instantiated on $O$ as $P(O)$ and $A(O)$, respectively, 
which is a union of Cartesian products of predicates/actions and their arguments, 
i.e., they represent the set of all ground propositions and actions.
A state $s\subseteq P(O)$ is a set of propositions that are true in that state.
An action is a 4-tuple $\brackets{\pre(a),\adde(a),\dele(a),\cost(a)}$,
where $\pre(a), \adde(a), \dele(a) \in P(X)$ are preconditions, add-effects, and delete-effects,
and $\cost(a) \in \R$ is a cost of taking the action $a$.
In this paper, we primarily assume a unit-cost domain where $\cost(a)=1$ for all $a$.
Given a \emph{current state} $s$,
a ground action $a_{\dagger}\in A(O)$ is \emph{applicable} when $\pre(a_{\dagger})\subseteq s$,
and applying an action $a_{\dagger}$ to $s$ yields a \emph{successor state}
$a_{\dagger}(s) = (s \setminus \dele(a_{\dagger})) \cup \adde(a_{\dagger})$.
Finally,
$I,G\subseteq P(O)$ are the initial state and a goal condition, respectively.
The task of classical planning is to find a \emph{plan} $(a_{\dagger}^1,\cdots,a_{\dagger}^n)$
which satisfies $a_{\dagger}^n \circ \cdots \circ a_{\dagger}^1(I) \supseteq G$
and every action $a_{\dagger}^i$ satisfies its preconditions at the time of using it.
The machine representation of a state $s$ and the goal condition $G$ is a bitvector of size $|P(O)|=\sum_{p\in P} O^{\ar{p}}$,
i.e., the $i$-th value of the vector is 1 when the corresponding $i$-th proposition is in $s$, or $G$.

\subsection{Markov Decision Processes}
\label{sec:mdp}

In general, RL methods address domains modeled as a discounted Markov decision
processes (MDP),
$\mathcal{M} = \brackets{\mathcal{S}, \mathcal{A}, T, r, q_0, \gamma}$ where
$\mathcal{S}$ is a set of states,
$\mathcal{A}$ is a set of actions,
$T(s,a,s'): \mathcal{S}\times \mathcal{A} \times \mathcal{S} \to \B$ encodes the probability $\Pr(s'|s,a)$
of transitioning from a state $s$ to a successor state $s'$ by an action $a$,
$r(s,a,s'): \mathcal{S}\times \mathcal{A} \times \mathcal{S} \to \R$ is a reward function,
$q_0$ is a probability distribution over initial states,
and $0\leq \gamma < 1$ is a discount factor.
In this paper, we restrict our attention to deterministic models because
PDDL domains are deterministic,
 and we have a deterministic mapping $T':\mathcal{S}\times \mathcal{A} \to \mathcal{S}$.
Given a \emph{policy} $\pi: \mathcal{S}\times \mathcal{A}\to \B$ representing a probability $\Pr(a|s)$
of performing an action $a$ in a state $s$,
we define a sequence of random variables $\braces{S_t}^\infty_{t=0}$, $\braces{A_t}^\infty_{t=0}$ and $\braces{R_t}^\infty_{t=0}$,
representing states, actions and rewards over time $t$.

Our goal is to find a policy maximizing its \emph{long term discounted cumulative rewards},
formally defined as a \emph{value function}
$V_{\gamma, \pi}(s) = \E_{A_t \sim \pi(S_t, \cdot)} \left[\sum_{t=0}^\infty \gamma^t R_t \mid S_0 = s \right].$
We also define an \emph{\qvalue} function to be the value of executing a given action and subsequently following some policy $\pi$, i.e.,
$Q_{\gamma, \pi}(s,a) = \E_{S_1\sim T(s,a,\cdot)} \left[R_0 + \gamma V_{\gamma,\pi}(S_1) \mid S_0 = s, A_0 = a \right].$
An \emph{optimal policy} $\pi^*$ is a policy that achieves the \emph{optimal value function} $V^*_\gamma=V_{\gamma,\pi^*}$
that satisfies $V^*_\gamma(s) \geq V_{\gamma,\pi}(s)$ for all states and policies.
$V^*_\gamma$ satisfies \emph{Bellman's equation}:
\begin{align}
V^*_\gamma(s) &= \max_{a \in \mathcal{A}} Q^*_\gamma(s, a) \;\;\; \forall s \in \mathcal{S}, \label{eq:bellman-opt}
\end{align}
where $Q^*_\gamma=Q_{\gamma,\pi^*}$ is referred to as the \emph{optimal \qvalue function}.
We may omit $\pi$ in $V_{\gamma,\pi}$, $Q_{\gamma,\pi}$ for clarity.

Finally, we can define a policy
by mapping \qvalues in each state to a probability distribution over actions.
For example, given an action-value function, $Q$, we can define a policy $\pi(s, a) = \softmax(Q(s, a)/\tau)$,
where $\tau > 0$ is a temperature that controls the greediness of the policy.
It returns a greedy policy $\argmax_a Q(s, a)$ when $\tau \to 0$; and approaches a uniform policy when $\tau \to \infty$.

\subsection{Formulating Classical Planning as an MDP}

There are two typical ways to formulate a classical planning problem
as an MDP.
In one strategy, given a transition $(s,a,s')$,
one may assign a reward of 1 when $s'\in G$, and 0 otherwise \citep{rivlin2020generalized}.
In another strategy, one may assign a reward of 0
when $s\in G$, and $-1$ otherwise (or, more generally $-\cost(a)$ in a non-unit-cost domain).
In this paper we use the second, \emph{negative-reward} model because it tends to induce more effective exploration in RL due to optimistic initial values~\citep{sutton2018reinforcement}.
Both cases are considered sparse reward problems because there is no
information about whether one action sequence is better than another
until a goal state is reached.

\section{Bridging Deep RL and AI Planning}
\label{sec:bridge}

We consider a multitask learning setting with a training time and a test time
 \citep{fern2011first}.
During training, classical planning problems from a single domain are available.
At test time, methods are evaluated on held-out problems from the same domain.
The transition model (in PDDL form) is known at both training and test time.

Learning to improve planning has been considered in RL. For example,
in AlphaGo \citep{alphago}, a value function was learned to provide
heuristic guidance to Monte Carlo Tree Search \citep{kocsis2006bandit}.
Applying RL techniques in our classical planning setting, however,
presents unique challenges.

\textbf{(P1): Preconditions and dead-ends.}
In MDPs, a failure to perform an action
is typically handled as a self-cycle to the current state
in order to guarantee that the state transition probability $T$ is well-defined for all states.
Another formulation
augments the state space with an absorbing state with a highly negative reward.
In contrast, classical planning does not handle non-deterministic outcomes (success and failure).
Instead, actions are forbidden at a state when its preconditions are not satisfied,
and a state is called a dead-end when no actions are applicable.
In a self-cycle formulation, random interaction with the environment could be inefficient
due to repeated attempts to perform inapplicable actions.
Also, the second formulation requires assigning an ad-hoc amount of negative reward to an absorbing state, which is not appealing.

\textbf{(P2): Objective functions.}
While the MDP framework itself does not necessarily assume discounting,
the majority of RL applications 
aim to maximize the expected cumulative \emph{discounted} rewards of trajectories.
In contrast, classical planning tries to minimize the sum of costs (negative rewards) along trajectories,
i.e., cumulative \emph{undiscounted} costs,
thus carrying the concepts in classical planning over to RL requires caution.

\textbf{(P3): Input representations.}
While much of the deep RL literature assumes an unstructured (e.g., images in Atari) or a factored input representation (e.g., location and velocity in cartpole),
classical planning deals with structured inputs based on FOL
to perform domain- and problem-independent planning.
This is problematic for typical neural networks, which assume a fixed-sized input.
Recently, several network architectures were proposed to achieve invariance to size and ordering,
i.e., neural networks for \emph{set}-like inputs \citep{zaheer2017deep}.
Graph/Hypergraph Neural Networks \citep[GNNs/HGNs]{scarselli2009graph,rivlin2020generalized,shen2020learning,ma2020online}
have also been recently used to encode FOL inputs.
While the choice of the architecture is arbitrary, our network should be able to handle FOL inputs.

\subsection{Value Iteration for Classical Planning}
\label{sec:vi}

Our main approach will be to learn a value function that can be used as a
heuristic to guide planning.
To learn estimated value functions, we build on the \emph{value iteration} (VI) algorithm (\refline{line:vi}, \refalgo{alg:vi}), where
a known model of the dynamics is used to incrementally update the estimates $V_{\gamma,\pi}(s)$ of
the optimal value function $V_{\gamma,\pi^*}(s)$.
The current estimates $V_{\gamma,\pi}(s)$ is updated by the r.h.s.\ of~\refeq{eq:bellman-opt} until a fixpoint is reached.
\begin{algorithm}[tb]
 \begin{algorithmic}[1]
  \STATE \emph{Value Iteration (VI)}:\label{line:vi}
  \WHILE{not converged}
  \FOR{$s\in \mathcal{S}$}\label{line:vi-inner-loop}
  \STATE $V_{\gamma,\pi}(s) \from \max_{a \in \mathcal{A}} Q_{\gamma,\pi}(s, a)$ \label{line:vi-update}
  \ENDFOR
  \ENDWHILE
  \STATE \vspace{4pt}\emph{Approximate RTDP with Replay Buffer}:\label{line:rtdp}
  \STATE Buffer $B\from \emptyset$ \label{line:replay1}
  \WHILE{not converged}
  \STATE $s\sim q_0$, $t\from 0$
  \WHILE{$t < D$ and $s$ is non-terminal}
  \STATE $a\from \argmax_a Q_{\gamma,\pi}(s,a)$
  \STATE $s\from T'(s,a)$
  \STATE $B\textit{.push}(s)$ \label{line:replay2}
  \STATE SGD($\frac{1}{2}\parens{V_{\gamma,\pi}(s) - \E_{a \in \mathcal{A}} Q_{\gamma,\pi}(s, a)}^2, B$) \label{line:sgd}
  \STATE $t\from t+1$
  \ENDWHILE
  \ENDWHILE
 \STATE \vspace{4pt}\emph{Approximate RTDP for Classical Planning}:\label{line:mrtdp}
 \STATE Buffer $B\from \red{[\emptyset, \emptyset, \emptyset, \ldots ]}$
 \WHILE{not converged}
 \STATE $\red{\brackets{\mathcal{D},O,I,G}}\sim q_0$, $t\from 0$, $s\from I$, \label{line:sampleinit}
 \WHILE{$t < D$, \red{$s\not\in G$}, $s$ is \red{not a deadlock}} \label{line:reset}
 \STATE $a\from \argmax_{a\in \red{\braces{a|\pre(a)\subseteq s}}} Q_{\gamma,\pi}(s,a)$ \label{line:action}
 \STATE $s\from T'(s,a)$
 \STATE $B\red{[|O|]}\textit{.push}(s)$ \label{line:buckets}
 \STATE SGD($\frac{1}{2}\parens{V_{\gamma,\pi}(s) - \E_{a \in \mathcal{A}} Q_{\gamma,\pi}(s, a)}^2, B$)
 \STATE $t\from t+1$
 \ENDWHILE
 \ENDWHILE
 \end{algorithmic}
 \caption{VI, RTDP, RTDP for Classical Planning}
 \label{alg:vi}
 \label{alg:rtdp}
 \label{algo:mrtdp}
\end{algorithm}
In classical planning, however,
state spaces are too large to enumerate its states (\refline{line:vi-inner-loop}),
or to represent the estimates $V_{\gamma,\pi}(s)$ in a tabular form (\refline{line:vi-update}).

To avoid the exhaustive enumeration of states in VI,
Real Time Dynamic Programming \citep[RTDP, \refline{line:rtdp}]{sutton2018reinforcement}
samples a subset of the state space based on the current policy.
In this work, we use \emph{on-policy} RTDP, which replaces the second $\max_a$ with $\E_a$ (\refline{line:sgd}) for the
current policy defined by the $\softmax$ of the current \qvalue estimates. On-policy methods
are known to be more stable but can sometimes lead to slower convergence.

Next, to avoid representing the value estimates in an exhaustive table,
we encode $V_{\gamma,\pi}$
using a neural network parameterized by weights $\theta$ and applying the Bellman updates approximately
with Stochastic Gradient Descent (\refline{line:sgd}).

As a common practice called \emph{experience replay} \citep{lin1993reinforcement,dqn},
we store the state history into a fixed-sized FIFO buffer $B$ (\reflines{line:replay1}{line:replay2}),
and update $V_{\gamma,\pi}(s)$ with mini-batches sampled from $B$ to leverage GPU parallelism.
The oldest record retires when $|B|$ reaches a limit.

We modify RTDP to address the assumptions \textbf{(P1)} in classical planning, resulting in \refline{line:mrtdp}.
First, in our multitask setting, where goals vary between problem instances, we wish to learn
a single goal-parameterized value function that generalizes across problems \citep{schaul2015universal}.
We omitted the goal for notational concision, but all of our
value functions are implicitly goal-parameterized, i.e., $V(s)=V(s,G)$.

Next, problem instances with different numbers of objects
have state representations (tensors) of varying sizes and dimensions.
Such a set of arrays with non-uniform shapes makes it challenging from
a mini-batch processing on GPUs.
Moreover, since larger problem instances typically require more steps to solve,
states from these problems are likely to dominate the replay buffer. This can make
updates to states from smaller problems rare, which can lead to catastrophic
forgetting. To address this, we separate the buffer into buckets (\refline{line:buckets}),
where states in one bucket are from problem instances with the same number of objects.
When we sample a mini-batch, we randomly select a bucket and randomly select states from this bucket.

Next, instead of terminating the inner loop and sampling the initial state in the same state space,
we redefine $q_0$ to be a distribution of problem instances,
and select a new training instance and start from its initial state (\refline{line:sampleinit}).

Finally, since $\argmax_a$ in RTDP is not possible at a state with no applicable actions (a.k.a. \emph{deadlock}),
we reset the environment upon entering such a state (\refline{line:reset}).
We also select actions only from applicable actions
and do not treat an inapplicable action as a self-cycle (\refline{line:action}).
Indeed,
training a value function along a trajectory that includes self-cycles has no benefit
because the test-time agents never execute them due to duplicate detection.

\subsection{Planning Heuristics as Dense Rewards}
\label{sec:reward-shaping}

The fundamental difficulty of applying RL-based approaches to
classical planning is the lack of dense reward to guide exploration.
We address this by combining heuristic functions (e.g., $\ff,\ad$)
with a technique called \emph{potential-based reward shaping}.
To correctly perform this technique, we should take care of
the difference between the discounted and non-discounted objectives \textbf{(P2)}.

Potential-based reward shaping \citep{ng1999policy} is a technique that helps RL algorithms
by modifying the reward function $r$.
Formally,
with a \emph{potential function} $\phi: \mathcal{S} \to \mathbb{R}$, a function of states,
we define a shaped reward function on
transitions, $\hat{r}: \mathcal{S} \times \mathcal{A} \times \mathcal{S} \to \mathbb{R}$, as follows:
\begin{align}
\hat{r}(s, a, s') &= r(s, a, s') + \gamma \phi(s') - \phi(s). \label{eq:shaping1}
\end{align}
Let $\mathcal{\hat{M}}$ be a MDP with a shaped reward $\hat{r}$, and $\mathcal{M}$ be the original MDP.
When the discount factor $\gamma<1$,
or when the MDP is \emph{proper},
i.e., every policy eventually ($t\to\infty$) reaches a terminal state with probability 1 under $\gamma=1$,
any optimal policy $\hat{\pi}^*$ of $\mathcal{\hat{M}}$ is an optimal policy $\pi^*$ of $\mathcal{M}$ regardless of $\phi$,
thus RL converges to an policy optimal in the original MDP $\mathcal{M}$.
Also, the optimal value function $\hat{V}^*_\gamma$ under $\mathcal{\hat{M}}$ satisfies
\begin{align}
V^*_\gamma(s) &= \hat{V}^*_\gamma(s) + \phi(s). \label{eq:shaping2}
\end{align}
In other words,
an agent trained in $\mathcal{\hat{M}}$
is learning an offset of the original optimal value function from the potential function.
The potential function thus acts as prior knowledge about the environment,
which initializes the value function to non-zero values \citep{wiewiora2003potential}.

Building on this theoretical background, we propose to leverage
existing domain-independent heuristics to define a potential function that
guides the agent while it learns to solve a given domain.
A naive approach that implements this idea is to define $\phi(s) = -h(s)$.
The $h$ value is negated because the MDP formulation seeks to \emph{maximize} reward
and $h$ is an estimate of cost-to-go, which should be minimized.
Note that the agent receives an additional reward
when $\gamma \phi(s') - \phi(s)$ is positive (\refeq{eq:shaping1}).
When $\phi=-h$, this means that approaching toward the goal and reducing $h$ is treated as a reward signal.
Effectively, this allows us to use a
domain-independent planning heuristic to generate dense rewards that
aid in the RL algorithm's exploration.

However,
this straightforward implementation has two issues:
\textbf{(1)} First, when the problem contains a dead-end,
the function may return $\infty$, i.e., $h: \mathcal{S} \to \mathbb{R}^{+0}\cup \braces{\infty}$.
This causes a numerical error in gradient-based optimization.
\textbf{(2)} Second,
the value function still requires a correction
even if $h$ is the ``perfect'' oracle heuristic $h^*$.
Recall that $V^*_\gamma$ is the optimal discounted value function with $-1$ rewards per step.
Given an optimal unit-cost cost-to-go $h^*(s)$ of a state $s$,
the discounted value function and the non-discounted cost-to-go can be associated as follows:
\begin{align}
V^*_\gamma(s) &= \sum_{t=1}^{h^*(s)} \gamma^t\cdot (-1) = -\frac{1-\gamma^{h^*(s)}}{1-\gamma} \not= -h^*(s).
\end{align}
Therefore, the amount of correction needed (i.e., $\hat{V}^*_\gamma(s) = V^*_\gamma(s) - \phi(s)$)
is not zero even in the presence of an oracle $\phi=-h^*$.
\emph{This is a direct consequence of discounting difference.}

To address these issues, we propose
to use the \emph{discounted value of the heuristic function} as a
potential function.
Recall that a heuristic function $h(s)$ is an estimate of the cost-to-go from
the current state $s$ to a goal.
Since $h(s)$ does not provide a \emph{concrete} idea of how to reach a goal,
we tend to treat it as a black box.
An important realization, however, is that
it nevertheless represents a sequence of actions;
thus its value can be \emph{decomposed into a sum of action costs} (below, left), and
we define a corresponding \emph{discounted heuristic function} $h_\gamma(s)$ (below, right):
\begin{align}
h(s) &= \sum_{t=1}^{h(s)} 1, & h_\gamma(s) &= \sum_{t=1}^{h(s)} \gamma^t\cdot 1 = \dfrac{1 - \gamma^{h(s)}}{1 - \gamma}.
\end{align}
Notice that
$\phi=-h^*_\gamma$ results in $\hat{V}^*_\gamma(s)=0$.
Also, $h_\gamma$ is bounded within $[0, \frac{1}{1-\gamma}]$, avoiding numerical issues.

\subsection{Value-Function Generalized over Problems}
\label{sec:nlm-based-rl}

To learn \emph{domain-dependent, instance-independent} heuristics,
the value function used in the reward-shaping framework discussed above
must be invariant to the number, the order, and the textual representation of propositions and objects in a PDDL definition \textbf{(P3)}.
We propose the use of Neural Logic Machines \cite[NLMs]{dong2018nlm},
a differentiable ILP system for a learning task over FOL inputs.
Below, we describe how it works and how we encode and pass states $s$ and goal condition $G$
to NLM to obtain $V(s,G)$.

\subsubsection{Neural Logic Machines}
\label{sec:nlm}

NLMs represent a state in terms of binary arrays representing the truth value of each proposition.
Propositions are grouped by the arity $N$ of the predicates they were grounded from.
This forms a set of $(N+1)$-d arrays denoted as $\vz/N$,
where the leading dimensions are indexed by objects and the last dimension is indexed by predicates of arity $N$.
For example, when we have objects \pddl{a}, \pddl{b}, \pddl{c} and four binary predicates \pddl{on}, \pddl{connected}, \pddl{above} and \pddl{larger},
we enumerate all combinations \pddl{on(a,a), on(a,b)} ... \pddl{larger(c,c)},
resulting in an array $\vz/2\in\B^{3\times 3 \times 4}$.
Similarly, we may have $\vz/1\in\B^{3\times 2}$ for 2 unary predicates,
and $\vz/3\in\B^{3\times 3 \times 3 \times 5}$ for 5 ternary predicates.
The total number of elements in all arrays combined matches the number of propositions
$|P(O)|=\sum_{p\in P} O^{\ar{p}}$.
In the following, we call this representation
a Multi-Arity Predicate Representation (MAPR).

NLMs are designed to learn a class of FOL rules with the following set of restrictions:
Every rule is a Horn rule,
no rule contains function terms (such as a function that returns an object),
there is no recursion,
and all rules are applied between neighboring arities.
Due to the lack of recursion, the set of rules can be stratified into layers.
Let $P_k$ be a set of intermediate conclusions in the $k$-th stratum.
The following set of rules are sufficient for
representing any rules in this class of rules \cite{dong2018nlm}:
\begin{align*}
 \text{(expand)}\quad      & \forall{x}_{\ar{\overline{p_{k}}}}; \overline{p_{k}}(X;x_{\ar{\overline{p_{k}}}}) \leftarrow p_{k}(X),                 \\
 \text{(reduce)}\quad      & \underline{p_{k}}(X) \leftarrow \exists x_{\ar{p_{k}}};p_{k}(X;x_{\ar{p_{k}}}),                   \\
 \text{(compose)}\quad     & p_{k+1}(X) \leftarrow \\
 & \mathcal{F}\parens{\bigcup_{\pi} \Bigg(\Big(P_{k}\cup \underline{P_{k}}\cup \overline{P_{k}})/\ar{p_{k+1}}\Big)\parens{\pi(X)}\Bigg)}.
\end{align*}
Here, $p_{k},\underline{p_{k}},\overline{p_{k}},p_{k+1}\in P_{k},\underline{P_{k}},\overline{P_{k}},P_{k+1}$ (respectively) are predicates,
$X=(x_1,\ldots)$ is a sequence of parameters,
and $\mathcal{F}(T)$ is a formula consisting of logical operations $\braces{\land,\lor,\lnot}$ and terms $T$.
Intermediate predicates $\underline{p_{k}}$ and $\overline{p_{k}}$ have one less / one more parameters than $p_{k}$,
e.g., when $\ar{p_{k}}=3$, $\ar{\overline{p_{k}}}=4$ and $\ar{\underline{p_{k}}}=2$.
$(P_k\cup \underline{P_k}\cup \overline{P_k})/{\ar{p_{k+1}}}$ extracts the predicates whose arity is the same as that of $p_{k+1}$.
$\pi(X)$ is a permutation of $X$, and
$\bigcup_\pi$ iterates over $\pi$ to generate propositional groundings with various argument orders.
$\mathcal{F}(\cdot)$ represents a formula that combines a subset of input propositions.
By chaining these set of rules from $P_k$ to $P_{k+1}$ for a sufficient number of times (e.g., from $P_1$ to $P_5$),
it is able to represent any FOL Horn rules without recursions \cite{dong2018nlm}.

\begin{figure}[tb]
 \centering
 \includegraphics[width=\linewidth]{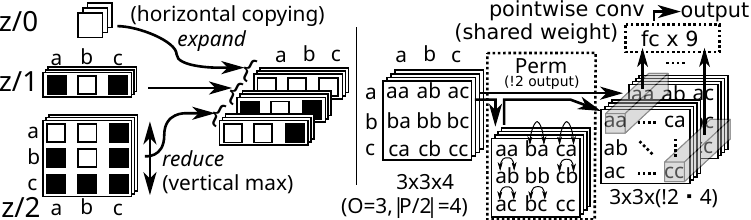}
 \caption{
\textbf{(Left)}
\function{expand} and \function{reduce} operations performed
on a boolean MAPR containing nullary, unary, and binary predicates
and three objects, \pddl{a}, \pddl{b}, and \pddl{c}.
Each white / black square represents a boolean value (true / false).
\textbf{(Right)}
\function{compose} operation performed on 4 binary predicates.
Each predicate is represented as a $3\times 3$ matrix,
resulting in a $\R^{3\times 3\times 4}$ tensor.
For a matrix, \function{perm} is equivalent to concatenating the matrix with its transposition,
resulting in a $\R^{3\times 3\times (!2\cdot 4)}$ tensor.
After \function{perm}, a shared fully-connected layer is applied to each combination of arguments
(such an operation is sometimes called a pointwise convolution).
}
 \label{nlm}
\end{figure}

All three operations (expand, reduce, and compose) can be implemented as tensor operations over MAPRs (\refig{nlm}).
Given a binary tensor $\vz/n$ of shape $O^n \times \setsize{P/n}$,
\emph{expand} copies the $n$-th axis to $n+1$-th axis resulting in a shape $O^{n+1} \times \setsize{P/n}$,
and
\emph{reduce} takes the $\max$ of $n$-th axis resulting in a shape $O^{n-1} \times \setsize{P/n}$,
representing $\exists$.

Finally, the \function{compose} operation combines the information between the neighboring tensors $\vz/n, \vz/_{n-1}, \vz/_{n+1}$.
In order to use the information in the neighboring arities ($P$, $\underline{P}$ and $\overline{P}$),
the input concatenates $\vz/n$ with $\function{expand}(\vz/_{n-1})$ and $\function{reduce}(\vz/_{n+1})$,
resulting in a shape $O^n \times C$ where $C=\setsize{P/n}+\setsize{P/_{n-1}}+\setsize{P/_{n+1}}$.
Next, a \function{Perm} function
enumerates and concatenates the results of permuting the first $n$ axes in the tensor,
resulting in a shape $O^n \times (!n\cdot C)$.
It then applies a $n$-D pointwise convolutional filter $f_n$ with $Q$ output features,
resulting in $O^n \times Q$, i.e.,
applying a fully connected layer to each vector of length $!n\cdot C$ while sharing the weights.
It is activated by any nonlinearity $\sigma$ to obtain the final result,
which is a sigmoid activation function in our implementation.
We denote the result as $\function{compose}(\vz, n, Q, \sigma)$.
Formally, $\forall j \in 1..n, \forall o_j \in 1..O$,
\begin{align*}
 &\Pi(\vz)=\function{perm}\Big(\function{expand}(\vz/_{n-1});\ \vz/n;\ \function{reduce}(\vz/_{n+1})\Big),
 \\
 &\function{compose}(\vz, n, Q, \sigma)_{o_1\cdots o_n}= \sigma(f_n(\Pi(\vz)_{o_1\cdots o_n})) \in \R^Q.
\end{align*}

An NLM contains $N$ (the maximum arity) \function{compose} operation for the neighboring arities,
with appropriately omitting both ends ($0$ and $N+1$) from the concatenation.
We denote the result as $\nlm_{Q,\sigma}(\vz)=(\function{compose}(\vz, 1, Q, \sigma),\cdots,\function{compose}(\vz, N, Q, \sigma))$.
These horizontal arity-wise compositions can be layered vertically, allowing the composition of
predicates whose arities differ more than 1 (e.g., two layers of NLM can combine unary and quaternary predicates).
Since $f_n$ is applied in a convolutional manner over $O^n$ object tuples,
the number of weights in an NLM layer
does not depend on the number of objects in the input.
However, it is still affected by the number of predicates in the input, which alters $C$.

\paragraph{NLMs/GNNs/HGNs}
We prefer NLMs over GNNs/HGNs for two reasons.
First, unlike GNNs/HGNs,
operations in NLMs have \emph{clear logical interpretations}: $\forall / \exists$ (\function{reduce} / \function{expand})
and combining input formula with different arguments (\function{compose}).
Next, 
the arity in NLMs' hidden layers can be \function{expand}-ed arbitrarily large.
GNNs are limited to binary/unary relations (edge/node-embeddings).
The arity of a hidden HGN layer can be higher, but must match the input.
FactorGNN \cite{zhang2020factor} is similar.

\subsubsection{Value Function as NLMs}

To represent a goal-generalized value function $V(s,G)$ with NLMs,
we concatenate each element of two sets of binary arrays:
One set representing the current state and another the goal conditions.
The last dimension of each array in the resulting set is twice larger.

When the predicates in the input PDDL domain have a maximum arity $N$,
we specify the maximum intermediate arity $M$ and the depth of NLM layers $L$ as a hyperparameter.
The intermediate NLM layers expand the arity up to $M$ using $\function{expand}$ operation,
and shrink the arity near the output because a value function has a scalar (arity 0) output.
For example, with $N=2$, $M=3$, $L=7$, the arity of each layer follows $(2,3,3,3,2,1,0)$.
Higher arities are not necessary near the output because
the information in each layer propagates only to the neighboring arities.
Since each expand/reduce operation only increments/decrements the arity by one,
$L,N,M$ must satisfy $N\leq M \leq L$.
Intermediate conclusions in NLM is fixed to $Q=8$.

The output of this NLM is unactivated, similar to a regression task,
because we use its raw value as the predicted correction to the heuristic function.
In addition, we implement NLM with a \emph{skip connection} that was popularized in ResNet image classification network \citep{he2016deep}:
The input of $l$-th layer is a concatenation of the outputs of all previous layers.
Due to the direct connections between the layers in various depths,
the layers near the input receive more gradient information from the output,
preventing the gradient vanishing problem in deep neural networks.

\section{Experimental Evaluation}
\label{sec:experiments}

Our objective is to see whether our RL agent can improve the efficiency of a
Greedy Best-First Search (GBFS), a standard algorithm for solving satisficing
planning problems, over a domain-independent heuristic. The efficiency
is measured by the number of node-evaluations performed during search.
We also place an emphasis on generalization: We hope that NLMs are able
to generalize from smaller training instances with fewer objects to instances
with more objects.

We train our RL agent with rewards shaped by $\ff$ and $\ad$ heuristics
obtained from \texttt{pyperplan} library.
We write blind heuristic $\forall s; \blind(s)=1$ to denote a baseline without shaping.
While our program is compatible with a wide range of unit-cost IPC domains
(see the list of 25 domains in \refappx{sec:feb16}),
we focus on extensively testing its selected subset with a large enough
number of independently trained models with different random seeds (20), to produce high-confidence results.
This is because RL algorithms tend to have
a large variance in their outcomes \citep{henderson2018deep},
induced by sensitivity to initialization, randomization in
exploration, and randomization in experience replay.

We trained our system on five domains in \cite{rivlin2020generalized}:
\pddl{4-ops blocksworld}, \pddl{ferry}, \pddl{gripper}, \pddl{logistics}, \pddl{satellite},
and three additional IPC domains: \pddl{miconic}, \pddl{parking}, and \pddl{visitall}.
In all domains, we generated
problem instances using existing
parameterized generators \citep{fawcett-et-al-icaps2011wspal}.
For each domain, we provided between 195 and 500 instances for training,
and between 250 and 700 instances for testing.
The generator parameters for test instances contain the ranges used for IPC instances
(See Appendix \reftbl{tbl:generator-parameters}).
We remove trivial instances whose initial states satisfy the goals,
which are produced by the generators occasionally,
especially for small parameter values used for training instances.
Each agent is trained for 50000 steps, which takes about 4 to 6 hours
on Xeon E5-2600 v4 and Tesla K80.
Hyperparameters can be found in \refappx{sec:hyper}.

\begin{table*}[tb]
\let\mc\multicolumn
\centering
\begin{adjustbox}{width=\linewidth}
\begin{tabular}{r|rrr|ccc||c||rr|rr|}
\toprule
 & \mc{3}{c|}{Baselines} & \mc{3}{c||}{Ours (mean\spm stderr (max) of 20 runs)} & GBFS & \mc{2}{c|}{GBFS} & \mc{2}{c|}{\spc{\small{GBFLS}\\\small{(incomparable)}}} \\
domain (total)  & $\blind$ & $\ad$ & $\ff$ & $\Hblind$                  & $\Had$                       & $\Hff$                    & -HGN & -H  & -V & (-H & -V) \\
\midrule
\pddl{blocks} (250)    & 0   & 126 & 87  & \textbf{73.1\spm 2.8(94)}           & \textbf{186.6\spm 7.5(229)}                                    & \textbf{104\spm 1.5(114)}                                      & 3  & 208 & 0 & 250 & 250 \\
\pddl{ferry} (250)     & 0   & 138 & 250 & \textbf{40.4\spm 3.2(62)}           & \textbf{233.9\spm 4(249)}                                      & {\scaleobj{0.4}{\textdbend}} \red{\textsf{{250\spm 0(250)}}}   & 27 & 240 & 0 & 250 & 250 \\
\pddl{gripper} (250)   & 0   & 250 & 250 & \textbf{47.5\spm 5(85)}             & {\scaleobj{0.4}{\textdbend}} \red{\textsf{{250\spm 0(250)}}}   & {\scaleobj{0.4}{\textdbend}} \red{\textsf{{250\spm 0(250)}}}   & 63 & 139 & 0 & 250 & 250 \\
\pddl{logistics} (250) & 0   & 106 & 243 & 0\spm 0(0)                          & \uline{54.1\spm 6.8}(\textbf{115})                             & \uline{79.8\spm 12.9(189)}                                     & -  & 0   & 0 & 30  & 33  \\
\pddl{miconic} (442)   & 171 & 442 & 442 & \uline{143.3\spm 6.8}(\textbf{246}) & {\scaleobj{0.4}{\textdbend}} \red{\textsf{442\spm 0(442)}}     & {\scaleobj{0.4}{\textdbend}} \red{\textsf{440.8\spm 1.3(442)}} & -  & 0   & 0 & 442 & 442 \\
\pddl{parking} (700)   & 0   & 607 & 700 & \textbf{0.9\spm 0.2(3)}             & \textbf{619\spm 32.4(689)}                                     & {\scaleobj{0.4}{\textdbend}} \red{\textsf{696.9\spm 0.5(700)}} & -  & 333 & 0 & 403 & 357 \\
\pddl{satellite} (250) & 0   & 249 & 222 & \textbf{26.5\spm 5(99)}             & {\scaleobj{0.4}{\textdbend}} \red{\textsf{233.3\spm 5.2(250)}} & 163.2\spm 11(205)                                              & -  & 9   & 0 & 137 & 135 \\
\pddl{visitall} (252)  & 252 & 252 & 252 & \uline{207.6\spm 5.3(238)}          & {\scaleobj{0.4}{\textdbend}} \red{\textsf{251.9\spm 0.1(252)}} & {\scaleobj{0.4}{\textdbend}} \red{\textsf{252\spm 0(252)}}     & -  & 101 & 0 & 249 & 249 \\
\bottomrule
\end{tabular}
\end{adjustbox}
\caption{
Coverage of GBFS with 100,000 node evaluations limit.
Our scores are highlighted in \textbf{bold} and \uline{underline} when
our average score is significantly better/worse than the baseline ($|\text{ours}-\text{baseline}|>\text{stderr}$, where $\text{stderr}=\frac{\text{stdev}}{\sqrt{20}}$).
Some scores are {\scaleobj{0.4}{\textdbend}} \red{\textsf{marked for caution}} when both ours and baselines solved nearly all instances.
They were too easy to measure coverage differences, and thus the lack thereof does not imply the lack of improvement in the heuristics.
Instead, node evaluation plots (\refig{fig:eval}) show reduction in the search effort.
In conclusion, our approach generally improves the performance in many domains
except logistics, where only the best seed of $\ad$ managed to improve upon the baseline (106 $\to$ 115).
}
\label{tab:scores}
\end{table*}

\begin{figure*}[tb]
\centering
\begin{minipage}{0.38\linewidth}
\begin{adjustbox}{width=\linewidth}
\begin{tabular}{l|rr|rr|rr|}
\toprule
 & $\Hblind$ & $\blind$ & $\Had$ & $\ad$ & $\Hff$ & $\ff$\\
\midrule
\pddl{blocks} & 94 & 0 & 224 & 5 & 109 & 9\\
\pddl{ferry} & 62 & 0 & 249 & 0 & 250 & 0\\
\pddl{gripper} & 85 & 0 & 250 & 0 & 200 & 50\\
\pddl{logistics} & 0 & 0 & 108 & 14 & 167 & 77\\
\pddl{miconic} & 234 & 17 & 381 & 61 & 442 & 0\\
\pddl{parking} & 2 & 0 & 508 & 105 & 484 & 173\\
\pddl{satellite} & 93 & 0 & 115 & 110 & 155 & 63\\
\pddl{visitall} & 192 & 60 & 216 & 36 & 192 & 59\\
\bottomrule
\end{tabular}
\end{adjustbox}
\end{minipage}
\hfill
\begin{minipage}{0.43\linewidth}
\includegraphics[width=0.24\linewidth]{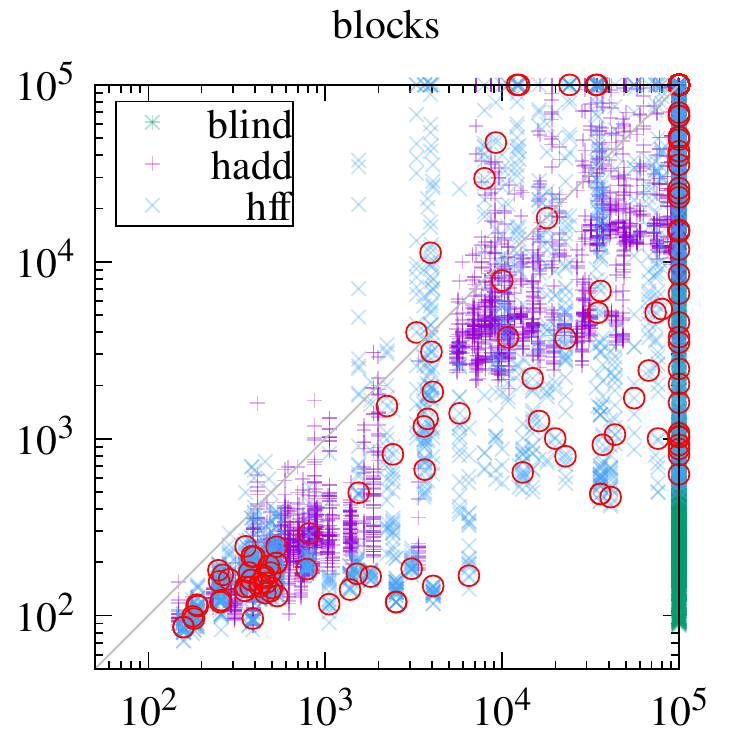}
\includegraphics[width=0.24\linewidth]{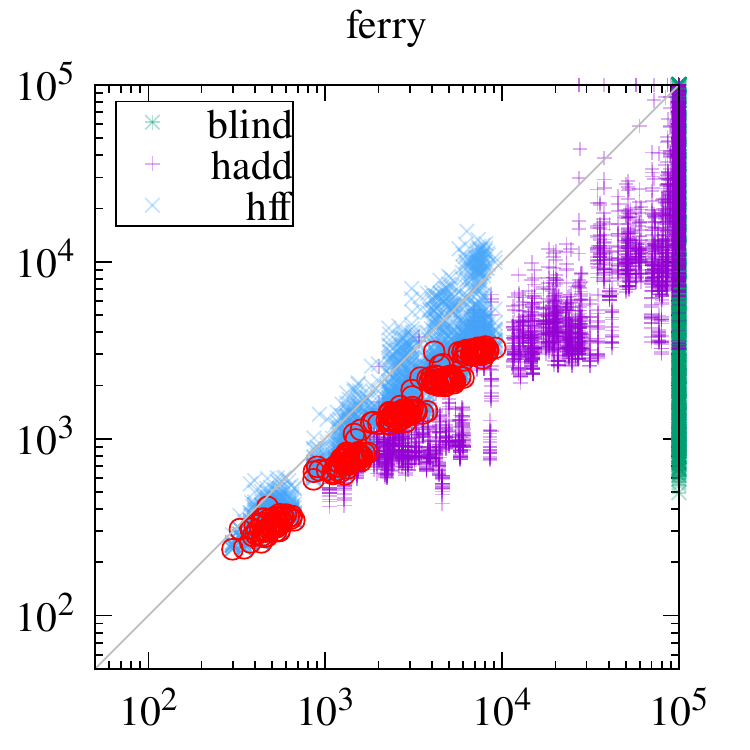}
\includegraphics[width=0.24\linewidth]{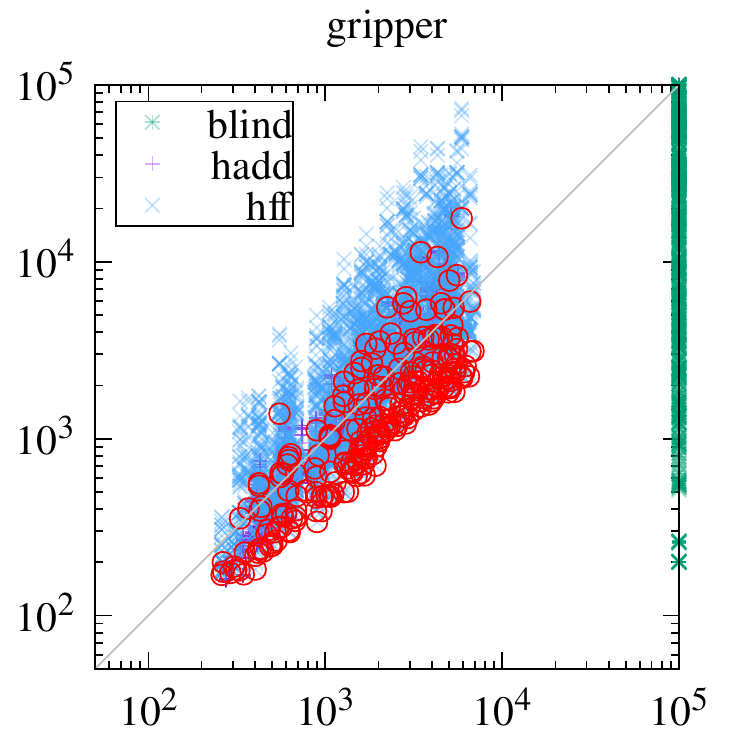}
\includegraphics[width=0.24\linewidth]{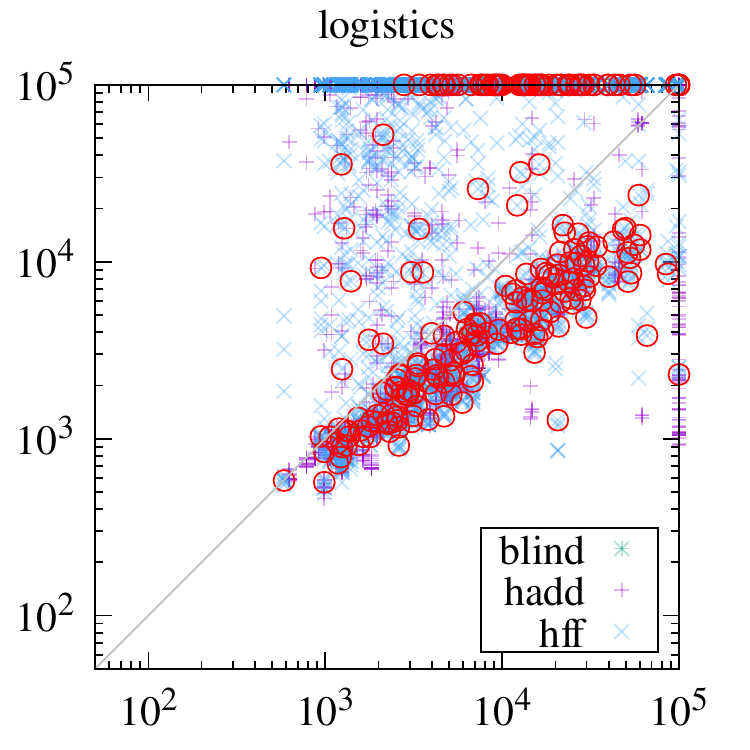}\\
\includegraphics[width=0.24\linewidth]{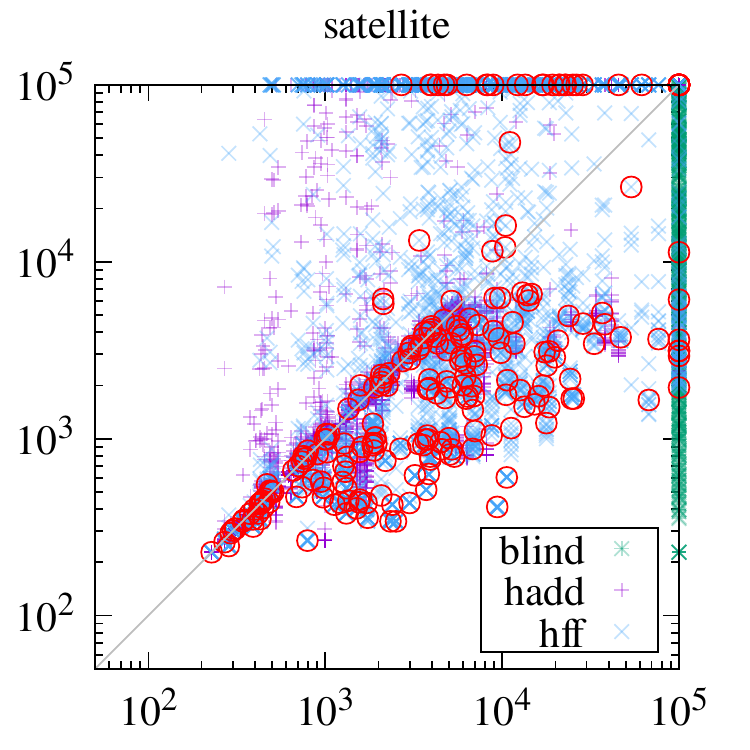}
\includegraphics[width=0.24\linewidth]{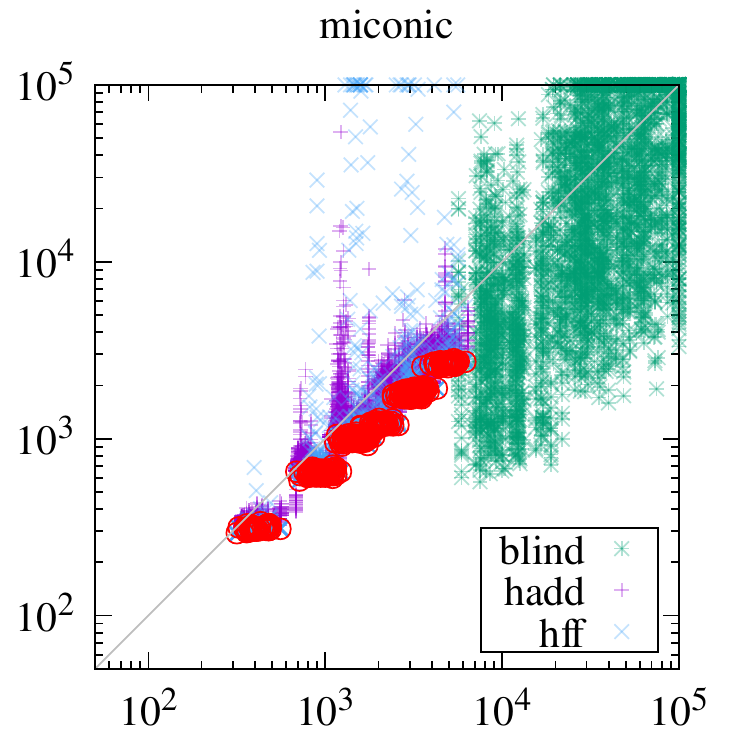}
\includegraphics[width=0.24\linewidth]{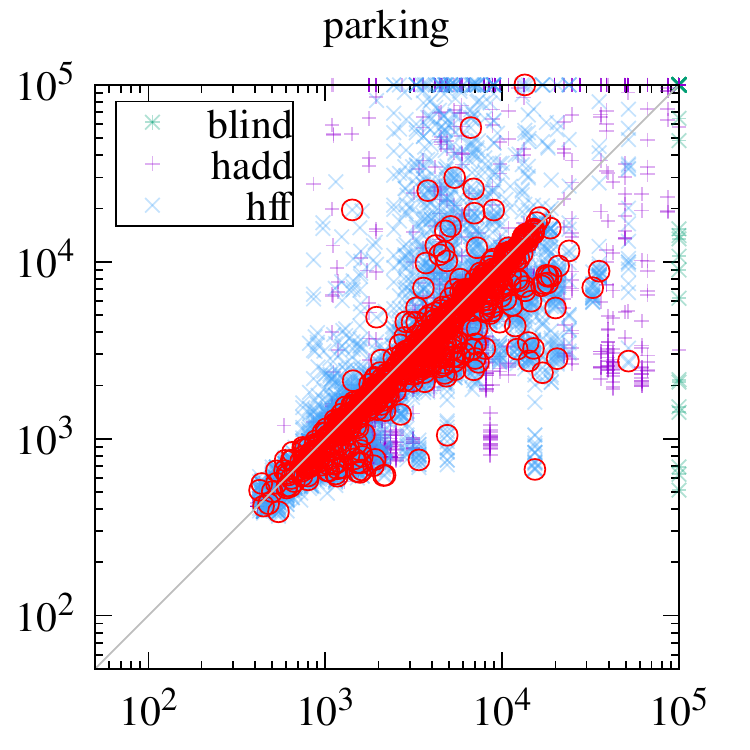}
\includegraphics[width=0.24\linewidth]{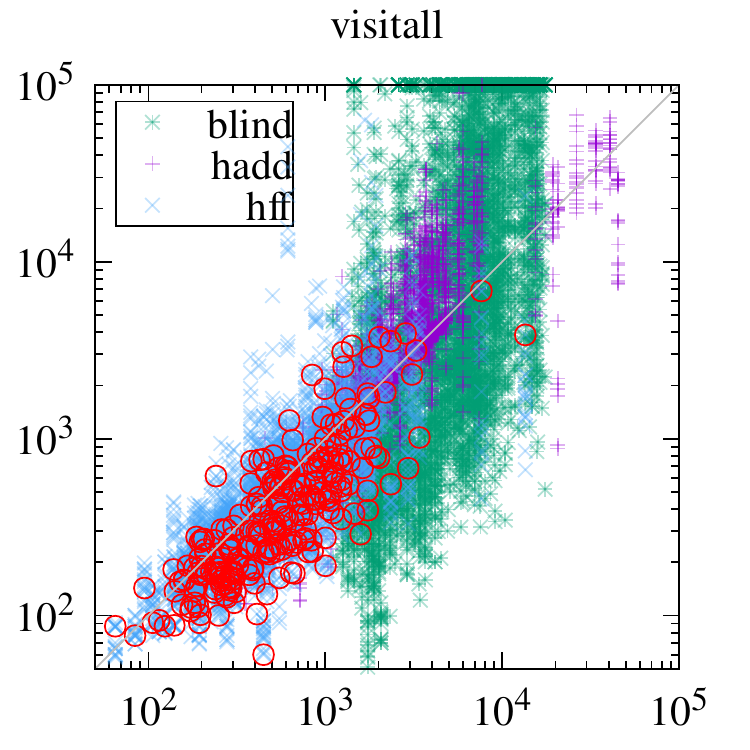}
\end{minipage}
\hfill
\begin{minipage}{0.14\linewidth}
\includegraphics[width=\linewidth]{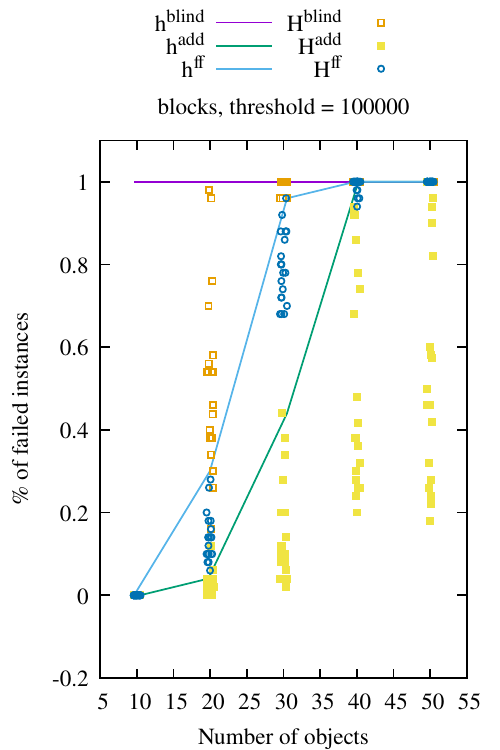}
\end{minipage}

\caption{
\textbf{Best viewed on computer screens.}
\textbf{(Left)}
Number of instances where $\function{GBFS}(H)$ had less evaluations than $\function{GBFS}(h)$ did (and vice versa),
excluding ties and instances failed by both.
$\function{GBFS}(H)$ is a result of the best training seed with the least sum of evaluations across the instances in a domain.
They show that RL improves the performance both with ($\Hff, \Had$) and without ($\Hblind$) reward shaping in the best case.
\textbf{(Middle)}
Scatter plot showing the number of node evaluations on 8 domains,
where $x$-axis is for GBFS with $\blind, \ff, \ad$ and $y$-axis is for $\Hblind, \Hff, \Had$.
Each point corresponds to a single test problem instance.
Results of 20 random seeds are plotted against a single deterministic baseline.
Failed instances are plotted on the border.
\red{\textbf{Red points}} highlight the best seed from $\Hff$.
\textbf{(Right)}
The rate of finding a solution ($y$-axis) for \pddl{blocks} instances and the number of objects ($x$-axis),
demonstrating that the improvements generalize to larger instances.
The agents are trained on 2-6 objects while the test instances contain 10-50 objects (\reftbl{tbl:generator-parameters}).
Results on other domains are in the appendix (\refig{fig:objs-full}).
}
\label{fig:eval}
\end{figure*}

We ran GBFS on the test instances
using $-V(s,G)=-\hat{V}(s,G)+h_\gamma(s)$ as a heuristics.
(Discounting does not affect the expansion order in GBFS and unit cost domains.)
Instead of setting time or memory limits, we limited the
maximum node evaluations in GBFS to 100,000. If a problem was solved within the limit,
the configuration gets the score 1 for that instance,
otherwise it gets 0. The sum of the scores for each domain is called the coverage in that domain.
\reftbl{tab:scores} shows the coverage in each of the tested domains, comparing
our configurations to the baselines, as well as to the prior work
(\refsec{sec:prior}). The baselines are denoted by
their heuristic (e.g., $\ff$ is the GBFS with $\ff$), while our heuristics,
obtained by a training with reward shaping $\phi=-h_\gamma$, are
denoted with a capital $H$ (e.g., $\Hff$).
Additionally, \refig{fig:eval} goes beyond the pure coverage and compares the node evaluations.
These results answer the following questions:

\textbf{(Q1)} Do our agents learn heuristic functions at all, i.e., is $\gbfs(\blind)<\gbfs(\Hblind)$ (\textbf{\green{green}} dots in \refig{fig:eval}),
 where $\gbfs(\blind)$ is similar to breadth-first search with duplicate detection,
and $\gbfs(\Hblind)$ is baseline RL without reward shaping?
With the exception of \pddl{visitall} and \pddl{miconic}, $\gbfs(\blind)$ could not solve any instances in the test set,
while using the heuristics learned without shaping ($\gbfs(\Hblind)$) significantly improved coverage in 5 of the 6 domains.

\textbf{(Q2)} Do they improve over the baselines they were initialized with, i.e., is $\gbfs(h)<\gbfs(H)$?
In domains where they did not solve every instances (\pddl{blocks}, \pddl{ferry}, \pddl{logistics}, \pddl{parking}, \pddl{satellite}),
\reftbl{tab:scores} suggests that the reward-shaping-based training has successfully improved the coverage
in \pddl{blocks}, \pddl{ferry}, \pddl{parking}.
Since the number of solved instances is not a useful metric
in domains where both configurations solved nearly all instances
(lack of coverage improvement does not imply lack of improvement in efficiency),
we next compare the number of node evaluations, which directly evaluates the search efficiency.
\refig{fig:eval} shows that the search effort tends to be reduced, especially on the best seed.
However, this is sensitive to the random seed, and the improvement is weak on \pddl{logistics}.
These results suggest that while RL can improve the planning efficiency,
we need several iterations of random experiments to achieve improvements due to the high randomness of RL.

\textbf{(Q3)} Do our agents with reward shaping outperform our agents without shaping?
According to \reftbl{tab:scores}, $\Hff$ and $\Had$ outperforms $\Hblind$.
Notice that $\ff$ and $\ad$ also outperform $\blind$.
This suggest that the informativeness of the base heuristic used for reward shaping affects the quality of the learned heuristic.
This matches the theoretical expectation: The potential function plays the role of domain knowledge that initializes the policy.

\textbf{(Q4)}
Did heuristics accelerate exploration during training and contribute to the improvement?
\reftbl{tbl:goals} shows the number of goals reached during training,
indicating that
reward shaping helps the agent receive real rewards at goals more often.
See Appendix \refigs{fig:goals1}{fig:goals2} for cumulative plots.

\textbf{(Q5)} Do the learned heuristics 
maintain their improvement in larger problem instances, i.e., do they generalize to more objects?
\refig{fig:eval} (Right) plots the number of objects ($x$-axis) and the ratio of success ($y$-axis) over \pddl{blocks} instances.
The agents are trained on 2-6 objects while evaluated on 10-50 objects.
It shows that the heuristic accuracy is improved in instances whose size far exceeds the training instances for $\blind, \ff, \ad$.
Due to space limitations,
plots for the remaining domains
are in Appendix, \refig{fig:objs-full}.

\begin{table}[tb]
\centering
\begin{tabular}{l|lll}
\toprule
domain           & $\blind$ & $\ad$ & $\ff$ \\
\midrule
\pddl{blocks}    & 362$\pm$42 & 527$\pm$58          & \textbf{621$\pm$31} \\
\pddl{ferry}     & 516$\pm$52 & \textbf{976$\pm$19} & 949$\pm$21          \\
\pddl{gripper}   & 275$\pm$33 & \textbf{673$\pm$18} & 600$\pm$20          \\
\pddl{logistics} & 93$\pm$21  & \textbf{502$\pm$33} & \textbf{488$\pm$32} \\
\pddl{miconic}   & 542$\pm$18 & 708$\pm$9           & \textbf{722$\pm$9}  \\
\pddl{parking}   & 400$\pm$51 & \textbf{809$\pm$24} & \textbf{814$\pm$27} \\
\pddl{satellite} & 212$\pm$37 & \textbf{658$\pm$34} & \textbf{654$\pm$26} \\
\pddl{visitall}  & 211$\pm$22 & 198$\pm$20          & \textbf{350$\pm$17} \\
\bottomrule
\end{tabular}
\caption{
The cumulative number of goal states the agent has reached
during training. The numbers are average and standard deviation over 20 seeds.
Best numbers among heuristics are highlighted in bold,
with ties equally highlighted when there are no statistically significant differences
between them under Wilcoxon's rank-sum test ($p\geq 0.05$).
The results indicate that reward shaping significantly accelerates the exploration compared to no shaping ($\blind$).
}
\label{tbl:goals}
\end{table}

\paragraph{Comparison with Previous Work}
\label{sec:prior}

Next, we compared our learned heuristics with two recent state-of-the-art
learned heuristics. The first approach, STRIPS-HGN \citep{shen2020learning}, is
a supervised learning method that can learn
domain-dependent or domain-independent heuristics depending on the dataset.
It uses hypergraph networks (HGN), a generalization of Graph Neural Networks (GNNs)
\citep{scarselli2009graph}.
The authors have provided us with
pre-trained weights for three domains: \pddl{gripper},
\pddl{ferry}, and \pddl{blocksworld} for the domain-dependent setting.
STRIPS-HGN was originally developed and evaluated for use with \astar,
for obtaining near-optimal plans.  Since we do not optimize plan
quality in this work, we instead use it with GBFS, which helps finding the goal more quickly possibly at the expense of plan quality
and is the same algorithm we use for our heuristics.
We recognize it may not have been designed for this scenario and that therefore may not best demonstrate its strengths.
It remains useful as a baseline point of comparison for our work.
We denote this variant GBFS-HGN.

The second approach we compare to is GBFS-GNN \citep{rivlin2020generalized},
an RL-based heuristic learning method that trains a GNN-based value function.
The authors use Proximal Policy Optimization \citep{schulman2017proximal},
a state of the art RL method that stabilizes the training by limiting the amount of policy change in each step
(the updated policy stays in the proximity of the previous policy).
The value function $V(s)$ is a GNN optionally equipped with attentions \cite{velivckovic2017graph,vaswani2017attention}.
In addition, the authors proposed to adjust $V(s)$ by the policy $\pi(a|s)$ and its entropy $H_\pi = \sum_a \pi(a|s)\log \pi(a|s)$.
The heuristic value of the successor state $s'=a(s)$ is given by $h(s')=\frac{\pi(a|s)V(s)}{1+H_\pi}$.
We call it an \emph{$H$-adjusted value function}.

The authors also proposed a variant of GBFS which launches a greedy informed local search after each expansion.
We distinguish their algorithmic improvement and the heuristics improvement by
naming their search algorithm as Greedy Best First Lookahead Search (GBFLS).
Our formal rendition of GBFLS can be found in \refappx{sec:gbfls}.

We counted the number of test instances that are solved by these approaches within 100,000 node evaluations.
In the case of GBFLS, the evaluations also include the nodes that appear during the lookahead.
We evaluated GBFS-HGN on the domains where pretrained weights are available.
For GBFS-GNN, we obtained the source code from the authors (private communication) and minimally modified it
to train on the same training instances that we used for our approach.
We evaluated 4 variants of GBFS-GNN: GBFS-H, GBFS-V, GBFLS-H, and GBFLS-V,
where ``H'' denotes $H$-adjusted value function, and ``V'' denotes the original value function.
Note that fair evaluation should compare our method with GBFS-H/V, not GBFLS-H/V.

\reftbl{tab:scores} shows the results.
We first observe that large part of the success of GBFS-GNN should be attributed to the lookahead extension of GBFS.
This is because the score is GBFLS-V $\gg$ GBFS-H $\gg$ GBFS-V, i.e., GBFLS-V performs very well even with a bad heuristics ($V(s)$).
While we report the coverage for both GBFLS-H/V and GBFS-H/V, the configurations that are comparable to our setting are GBFS-H/V.
First, note that GBFS-HGN is significantly outperformed by all other methods.
Comparing to the other two, both $\Had$ and $\Hff$ outperform GBFS-H in 7 out of the 8 domains, losing only on \pddl{blocks}.
It is worth noting that $\Hblind$ outperforms GBFS-H in \pddl{miconic}, \pddl{satellite}, and \pddl{visitall}.
Since both $\Hblind$ and GBFS-H are trained without reward shaping,
the difference is due to the network shape (NLM vs GNN) and the training (Modified RTDP vs PPO).

\section{Related Work}

Early attempts to learn heuristics include
shallow, fully connected neural networks \citep{ArfaeeZH11},
its online version \citep{thayer2011learning},
combining SVMs \citep{cortes1995support} and NNs \citep{SatzgerK13},
learning a residual from heuristics \citep{yoon2008learning},
or learning a relative ranking between states \citep{garrett2016learning}.
More recently,
\citet{ferber-et-al-ecai2020} tested fully-connected layers in modern frameworks.
ASNet \citep{toyer2018action} learns domain-dependent heuristics using a GNN-like network.
They are based on supervised learning methods that require
the high-quality training dataset (accurate goal distance estimates of states) that are prepared separately.
Our RL-based approaches explore the environment by itself to collect data,
which is automated (pros) but could be sample-inefficient (cons).

A large body of work utilize ILP techniques to learn
a value function \cite{DBLP:conf/uai/GrettonT04}
features \cite{wu2010automatic},
pruning rules \cite{DBLP:conf/ecai/Krajnansky0BF14}, or
a policy function by classifying the best action \cite{fern2006approximate}.
NLM is a differentiable ILP system that subsumes first-order decision lists / trees used in ILP.

Other RL-based approaches include
Policy Gradient with FF to accelerate exploration for probabilistic PDDL \citep{buffet2007ff},
and PPO-based Meta-RL \citep{duanscbsa16} for PDDL3.1 discrete-continuous hybrid domains \citep{gutierrez2021meta}.
They do not use reward shaping, thus our contributions are orthogonal.

\citet{grounds2005combining} combined RL and STRIPS planning
with reward shaping, but in a significantly different setting:
They treat a 2D navigation as a two-tier hierarchical problem where
unmodified FF \citep{hoffmann01} or Fast Downward \citep{Helmert2006} are used as high-level planner,
then their plans are used to shape the rewards for the low-level RL agent.
They do not train the high-level planner.

\section{Conclusion}

In this paper, we proposed a domain-independent reinforcement learning framework for learning
domain-specific heuristic functions.
Unlike existing work on applying policy gradient to planning \citep{rivlin2020generalized},
we based our algorithm on value iteration.
We addressed the difficulty of training an RL agent with sparse rewards
using a novel reward-shaping technique which leverages existing heuristics developed in the literature.
We showed that our framework not only learns a heuristic function from scratch ($\Hblind$),
but also learns better if aided by heuristic functions (reward shaping).
Furthermore, the learned heuristics keeps outperforming the baseline over a wide range of problem sizes,
demonstrating its generalization over the number of objects in the environment.

\section*{Acknowledgment}

We thank \citeauthor{shen2020learning} and \citeauthor{rivlin2020generalized}
for kindly providing us their code.

\fontsize{9.0pt}{10.0pt}\selectfont

\clearpage
\appendix

\section{Appendix}

\subsection{Domain-Independent Heuristics for Classical Planning}
\label{sec:heuristics-detail}

In this section, we discuss various approximations of delete-relaxed optimal cost $h^+(s)$.
Given a classical planning problem $\brackets{P,O,A,I,G}$, and a state $s$,
each heuristics is typically implicitly conditioned by the goal condition.
$\ad$ heuristics is recursively defined as follows:

\begin{align}
 h^\text{add}(s,G) = \sum_{p\in G}
 \left\{
  \begin{array}{l}
   0\ \text{if}\ p\in s.\ \text{Otherwise,}\\
   \min_{\braces{a\in A\mid p\in\adde(a)}} \\
    \qquad \left[\cost(a)+\ad(s, \pre(a))\right]. \\
  \end{array}
 \right.
\end{align}

$\ff$ heuristics can be defined based on $\ad$ as a subprocedure.
The action $a$ which minimizes the second case ($p\not\in S$) of each of the definition above
is conceptually a ``cheapest action that achieves a subgoal $p$ for the first time'',
which is called a \emph{cheapest achiever} / \emph{best supporter} $\text{bs}(p,s)$ of $p$.
Using $h$ and its best supporter function, $\ff$ is defined as follows:

\begin{align}
 h^\text{FF}(s,G) &= \sum_{a\in \text{RelaxedPlan}(s,G)} \cost(a)\\
 \text{RelaxedPlan}(s,G) &= \bigcup_{p\in G}
 \left\{
  \begin{array}{l}
   \emptyset\ \text{if}\ p\in s.\ \text{Otherwise,}\\
   \braces{a} \cup \text{RelaxedPlan}(s,\pre(a)) \\
   \qquad \text{where}\ a=\text{bs}(p,s).
  \end{array}
 \right.\\
 \text{bs}(p,s)&=\argmin_{\braces{a\in A\mid p\in \adde(a)}} \left[\cost(a)+\ad(s, \pre(a))\right].
\end{align}

\subsection{Greedy Best First Search and Greedy Best First Lookahead Search \cite{rivlin2020generalized}}
\label{sec:gbfs}
\label{sec:gbfls}

Given a classical planning problem $\brackets{P,O,A,I,G}$,
we define its state space as a directed graph $(V,E)$
where $V=2^{P(O)}$, i.e., a power set of subsets of propositions $P(O)$.
Greedy Best First Search is a greedy version of \astar algorithm \citep{hart1968formal}, therefore we define \astar first.

We follow the optimized version of the algorithm discussed in \citet{burns2012implementing}
which does not use CLOSE list and avoids moving elements between CLOSE list and OPEN list
by instead managing a flag for each search state.
Let $f(s)=g(s)+h(s)$ be a function that computes the sum of $h(s)$ and $g(s)$,
where $g(s)$ is a value stored for each state which represents the currently known upper bound of the shortest path cost from the initial state $I$.
For every state $s$, $g(s)$ is initialized to infinity except $g(I)=0$.
Whenever a state $s$ is expended, $f(s)$ is a lower bound of the path cost that goes through $s$.
\astar algorithm is defined as in \refalgo{alg:astar}.
We simplified some aspects such as updating the parent node pointer, the rules for tiebreaking,
 or extraction of the plan by backtracking the parent pointers.
Notice that the update rule for $g$-values in \astar is a Bellman update specialized for a positive cost function.

\def\open{\text{OPEN}}
\def\close{\text{CLOSE}}
\begin{algorithm}[tb]
 \begin{algorithmic}[1]
  \STATE Priority queue $\open\from \emptyset$.
  \STATE $g(s) \from \infty$ for all $s \in V$ except $g(I)\from 0$.
  \STATE $s$.closed $\from \bot$ for all $s \in V$.
  \STATE $\open.\function{push}(I,f(I))$
  \WHILE{$\open\not=\emptyset$}
  \STATE State $s\from \open.\function{popmin}()$ \COMMENT{Expansion}
  \IF{$s\supseteq G$} \label{list:astar-goal}
  \RETURN $s$
  \ENDIF
  \IF{$s$.closed $= \top$}
  \STATE \textbf{continue}
  \ELSE
  \STATE $s$.closed $\from \top$
  \ENDIF
  \FOR{$a\in \braces{a\in A\mid \pre(a)\subseteq s}$ }
  \STATE successor $t=a(s)$ \COMMENT{Note: Appropriate caching of $t$ is necessary.}
  \IF{$g(t) > \cost(a)+g(s)$}
  \STATE $g(t)\from \cost(a)+g(s)$   \COMMENT{Bellman update}
  \STATE $t$.closed $\from \bot$     \COMMENT{Reopening}
  \STATE $\open.\function{push}(t,f(t))$
  \ENDIF
  \ENDFOR
  \ENDWHILE
 \end{algorithmic}
\caption{\astar algorithm for a planning problem $\brackets{P,O,A,I,G}$ with state space $(V,E)$.}
\label{alg:astar}
\end{algorithm}

Three algorithms can be derived from \astar by redefining the sorting key $f(s)$ for the priority queue $\open$.
First, ignoring the heuristic function by redefining $f(s)=g(s)$ yields Dijkstra's search algorithm.
Another is weighted \astar \citep{pohl1970heuristic},
where we redefine $f(s)=g(s)+w\cdot h(s)$ for some value $w\geq 1$,
which results in trusting the heuristic guidance relatively more greedily.

As the extreme version of WA*, conceptually $w\to \infty$ yields
the Greedy Best First Search algorithm \cite{russell1995artificial} which completely greedily trusts the heuristic guidance.
In practice, we implement it by ignoring the $g(s)$ value, i.e., $f(s)=h(s)$.
This also simplifies some of the conditionals in \refalgo{alg:astar}:
There is no need for updating the $g$ value, or reopening the node.
In addition, purely satisficing algorithm like GBFS can enjoy an additional enhancement called \emph{early goal detection}.
In \astar, the goal condition is checked when the node is popped from the OPEN list (\refline{list:astar-goal}) --
if we detect the goal early, it leaves the possibility that it returns a suboptimal goal node.
In contrast,
since we do not have this optimality requirement in GBFS,
the goal condition can be checked in \refline{list:gbfs-goal} where the successor state is generated.
GBFS is thus defined as in \refalgo{alg:gbfs}.

\begin{algorithm}[tb]
 \begin{algorithmic}[1]
  \STATE Priority queue $\open\from \emptyset$.
  \STATE $s$.closed $\from \bot$ for all $s \in V$.
  \STATE $\open.\function{push}(I,h(I))$
  \WHILE{$\open\not=\emptyset$}
  \STATE State $s\from \open.\function{popmin}()$ \COMMENT{Expansion}
  \IF{$s$.closed $= \top$}
  \STATE \textbf{continue}
  \ELSE
  \STATE $s$.closed $\from \top$
  \ENDIF
  \FOR{$a\in \braces{a\in A\mid \pre(a)\subseteq s}$ }
  \STATE successor $t=a(s)$ \COMMENT{Note: Appropriate caching of $t$ is necessary.}
  \IF{$t\supseteq G$} \label{list:gbfs-goal}
  \RETURN $t$ \COMMENT{Early goal detection.}
  \ENDIF
  \STATE $\open.\function{push}(t,h(t))$
  \ENDFOR
  \ENDWHILE
 \end{algorithmic}
\caption{GBFS algorithm for a planning problem $\brackets{P,O,A,I,G}$ with state space $(V,E)$.}
\label{alg:gbfs}
\end{algorithm}

Finally,
\citet{rivlin2020generalized} proposed an unnamed extension of GBFS which performs a depth-first lookahead after a node is expanded.
We call the search algorithm Greedy Best First Lookahead Search (GBFLS), defined in \refalgo{alg:gbfls}.
We perform the same early goal checking during the lookahead steps.
Note that the nodes are added only when the current node is expanded;
Nodes that appear during the lookahead are not added to the OPEN list.
However, these nodes must be counted as evaluated node
because it is subject to goal checking and because we evaluate their heuristic values.
The lookahead has an artificial depth limit $D$ which is defined as $D=5\cdot\ff(I)$,
i.e., 5 times the value of the FF heuristics at the initial state.
When $\ff(I)=\infty$, the limit is set to 50, according to their code base.

\begin{algorithm}[tb]
 \begin{algorithmic}[1]
  \STATE Priority queue $\open\from \emptyset$.
  \STATE $s$.closed $\from \bot$ for all $s \in V$.
  \STATE $\open.\function{push}(I,h(I))$
  \WHILE{$\open\not=\emptyset$}
  \STATE State $s\from \open.\function{popmin}()$ \COMMENT{Expansion}
  \IF{$s$.closed $= \top$}
  \STATE \textbf{continue}
  \ELSE
  \STATE $s$.closed $\from \top$
  \ENDIF
  \FOR{depth $d<D$}
  \FOR{$a\in \braces{a\in A\mid \pre(a)\subseteq s}$ }
  \STATE successor $t=a(s)$ \COMMENT{Note: Appropriate caching of $t$ is necessary.}
  \IF{$t\supseteq G$}
  \RETURN $t$ \COMMENT{Early goal detection.}
  \ENDIF
  \STATE $\open.\function{push}(t,h(t))$ \textbf{if} $d=0$
  \ENDFOR
  \STATE $a\gets \argmin_{a} h(a(s))$
  \STATE $s\gets a(s)$
  \ENDFOR
  \ENDWHILE
 \end{algorithmic}
\caption{GBFLS algorithm for a planning problem $\brackets{P,O,A,I,G}$ with state space $(V,E)$.}
\label{alg:gbfls}
\end{algorithm}

\subsection{Implementation}

Our implementation combines the \texttt{jax} auto-differentiation
framework for neural networks, and
\texttt{pyperplan} for parsing and to obtain the heuristic value of $\ff$ and $\ad$.

\subsection{Generator Parameters}

\reftbl{tbl:generator-parameters} contains a list of parameters used to generate the training and testing instances.
Since generators have a tendency to create an identical instance especially in smaller parameters,
we removed the duplicates by checking the md5 hash value of each file.

\begin{table*}[htb]
\centering
 \begin{tabular}{|c|c|c|}
Domain           & Parameters                                                                             & $|O|$\\
blocks/train/    & 2-6 blocks x 50 seeds                                                                  & 2-6\\
blocks/test/     & 10,20,..,50 blocks x 50 seeds                                                          & 10-50\\
ferry/train/     & 2-6 locations x 2-6 cars x 50 seeds                                                    & 4-7\\
ferry/test/      & 10,15,...30 locations and cars x 50 seeds                                              & 20-60\\
gripper/train/   & 2,4...,10 balls x 50 seeds (initial/goal locations are randomized)                     & 6-14\\
gripper/test/    & 20,40,...,60 balls x 50 seeds (initial/goal locations are randomized)                  & 24-64\\
logistics/train/ & 1-3 airplanes x 1-3 cities x 1-3 city size x 1-3 packages x 10 seeds                   & 5-13\\
logistics/test/  & 4-8 airplanes/cities/city size/packages x 50 seeds                                     & 32-96\\
satellite/train/ & 1-3 satellites x 1-3 instruments x 1-3 modes x 1-3 targets x 1-3 observations          & 15-39\\
satellite/test/  & 4-8 satellites/instruments/modes/targets/observations x 50 seeds                       & 69-246\\
miconic/train/   & 2-4 floors x 2-4 passengers x 50 seeds                                                 & 8-12\\
miconic/test/    & 10,20,30 floors x 10,20,30 passengers x 50 seeds                                       & 24-64 \\
parking/train/   & 2-6 curbs x 2-6 cars x 50 seeds                                                        & 8-16\\
parking/test/    & 10,15,..,25 curbs x 10,15,..25 cars x 50 seeds                                         & 24-54\\
visitall/train/  & For $n\in 3..5$, $n$x$n$ grids, 0.5 or 1.0 goal ratio, $n$ blocked locations, 50 seeds & 8-22\\
visitall/test/   & For $n\in 6..8$, $n$x$n$ grids, 0.5 or 1.0 goal ratio, $n$ blocked locations, 50 seeds & 32-58\\
 \end{tabular}
\caption{List of parameters used for generating the training and testing instances.}
\label{tbl:generator-parameters}
\end{table*}

\subsection{Hyperparameters}
\label{sec:hyper}

We trained our network with a following set of hyperparameters:
Maximum episode length $D=40$,
Learning rate 0.001,
discount rate $\gamma=0.999999$,
maximum intermediate arity $M=3$,
number of layers $L=4$ in \pddl{satellite} and \pddl{logistics},
while $L=6$ in all other domains,
the number of features in each NLM layer $Q=8$,
batch size 25,
temperature $\tau=1.0$ for a policy function (\refsec{sec:mdp}),
replay buffer size $|B|=6000$,
and the total number of SGD steps to 50000, which determines the length of the training.
We used $L=4$ for those two domains to address GPU memory usage:
Due to the size of the intermediate layer $O^n\times (n!\cdot C)$, NLM sometimes requires a large amount of GPU memory.
Each training takes about 4 to 6 hours, depending on the domain.

\subsection{Preliminary Results on Compatible Domains}
\label{sec:feb16}

We performed a preliminary test on a variety of IPC classical domains that are supported by our implementation.
The following domains worked without errors:
\pddl{barman-opt11-strips}, \pddl{blocks}, \pddl{depot}, \pddl{driverlog}, \pddl{elevators-opt11+sat11-strips},
\pddl{ferry}, \pddl{floortile-opt11-strips}, \pddl{freecell}, \pddl{gripper}, \pddl{hanoi}, \pddl{logistics00},
\pddl{miconic}, \pddl{mystery}, \pddl{nomystery-opt11-strips}, \pddl{parking-opt11+sat11-strips},
\pddl{pegsol-opt11-strips}, \pddl{pipesworld-notankage}, \pddl{pipesworld-tankage}, \pddl{rovers},
\pddl{satellite}, \pddl{scanalyzer-08-strips}, \pddl{sokoban-opt11-strips}, \pddl{tpp},
\pddl{transport-opt11+sat08-strips}, \pddl{visitall-opt11-strips}, \pddl{zenotravel}.

\subsection{Full Results}

\begin{figure*}[tb]
\includegraphics[width=0.24\linewidth]{img___static___objs-nov9-blocks-_-gbfs-100000.pdf}
\includegraphics[width=0.24\linewidth]{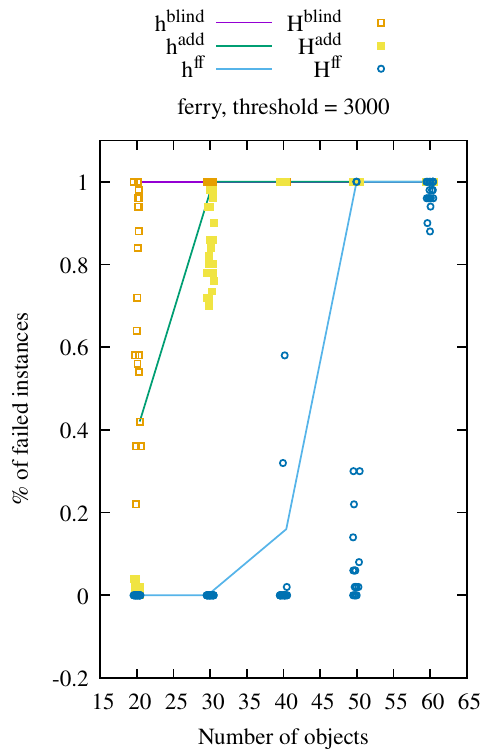}
\includegraphics[width=0.24\linewidth]{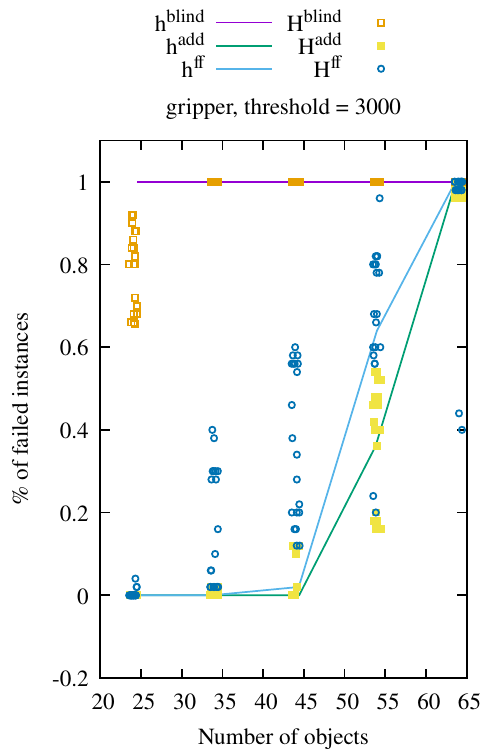}
\includegraphics[width=0.24\linewidth]{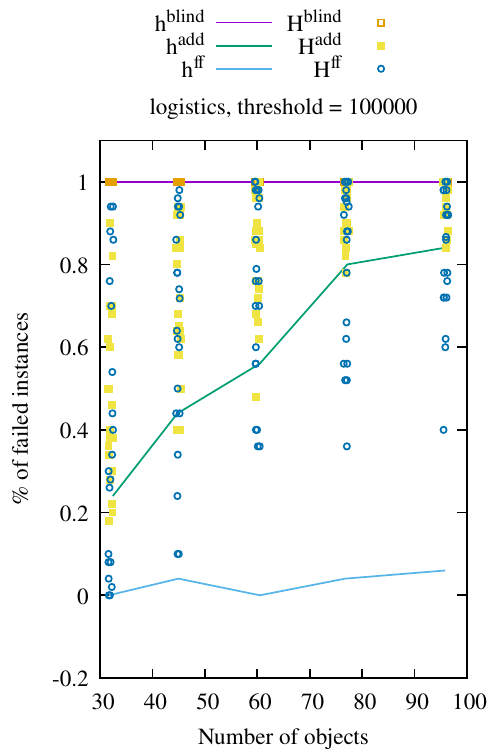}\\
\includegraphics[width=0.24\linewidth]{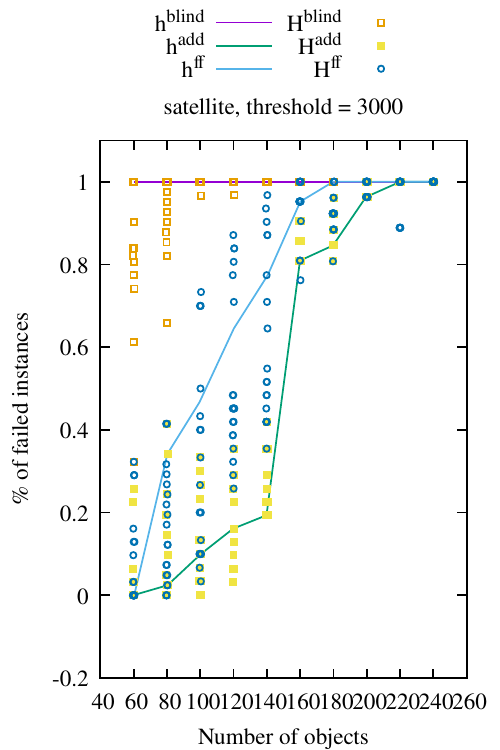}
\includegraphics[width=0.24\linewidth]{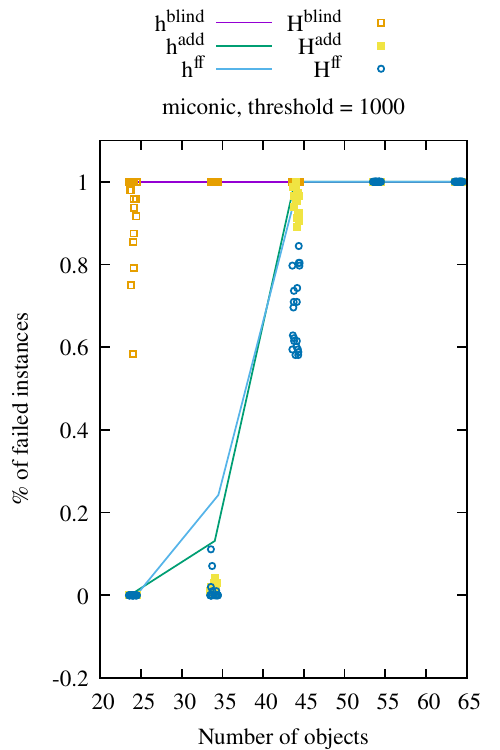}
\includegraphics[width=0.24\linewidth]{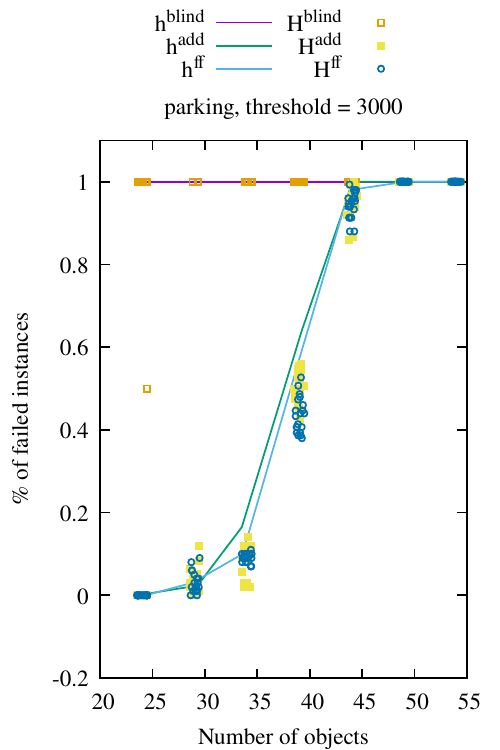}
\includegraphics[width=0.24\linewidth]{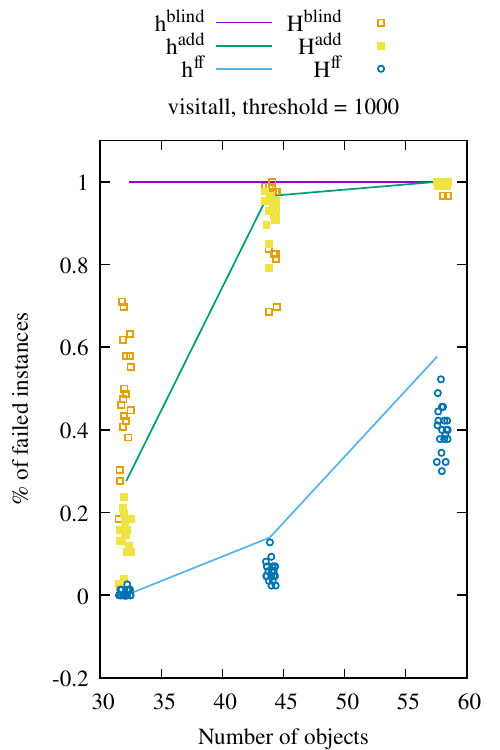}
\caption{
The rate of successfully finding a solution ($y$-axis) for instances with a certain number of objects ($x$-axis).
Learned heuristic functions outperform their original baselines used for reward shaping in most domains.
Since the initial maximum node evaluation is too permissive,
we manually set a threshold for the number of node evaluations for each domain and
filtered the instances when the node evaluation exceeded this threshold.
This filtering emphasizes the difference because
both the learned and the baseline variants may have solved all instances.
}
\label{fig:objs-full}
\end{figure*}

\refig{fig:objs-full} contains the full results of \refig{fig:eval} (Right).

\begin{figure*}[tb]
\includegraphics[width=0.33\linewidth]{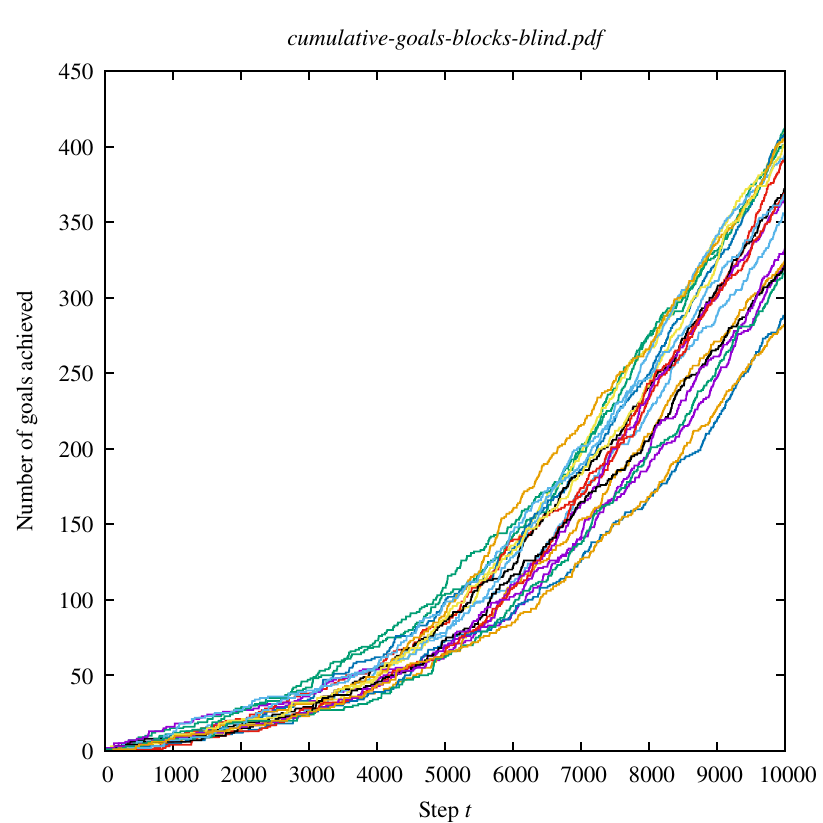}
\includegraphics[width=0.33\linewidth]{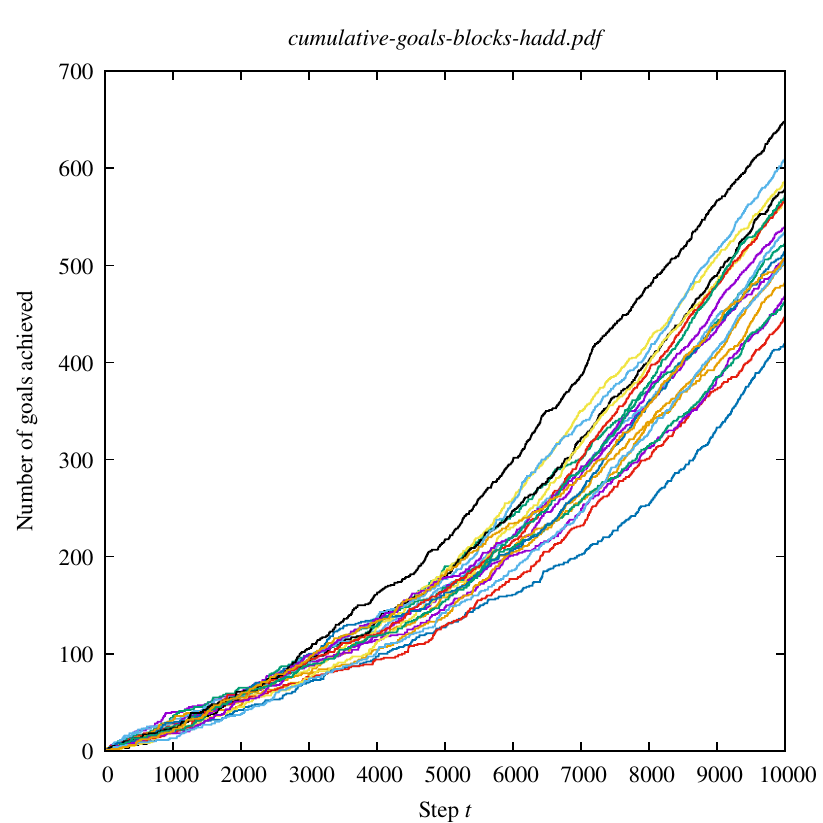}
\includegraphics[width=0.33\linewidth]{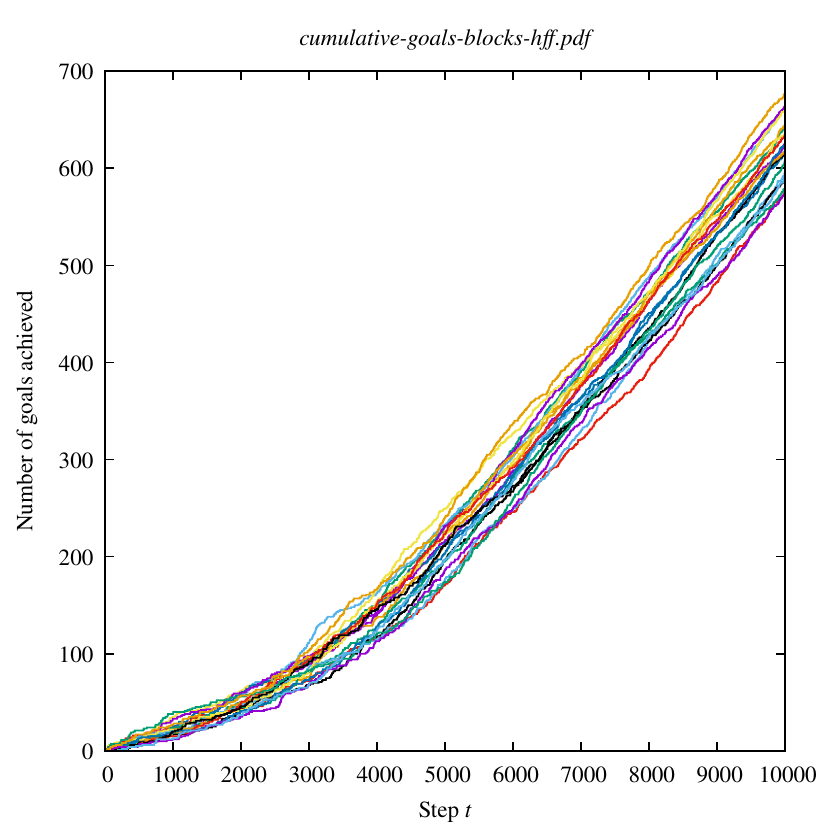}
\includegraphics[width=0.33\linewidth]{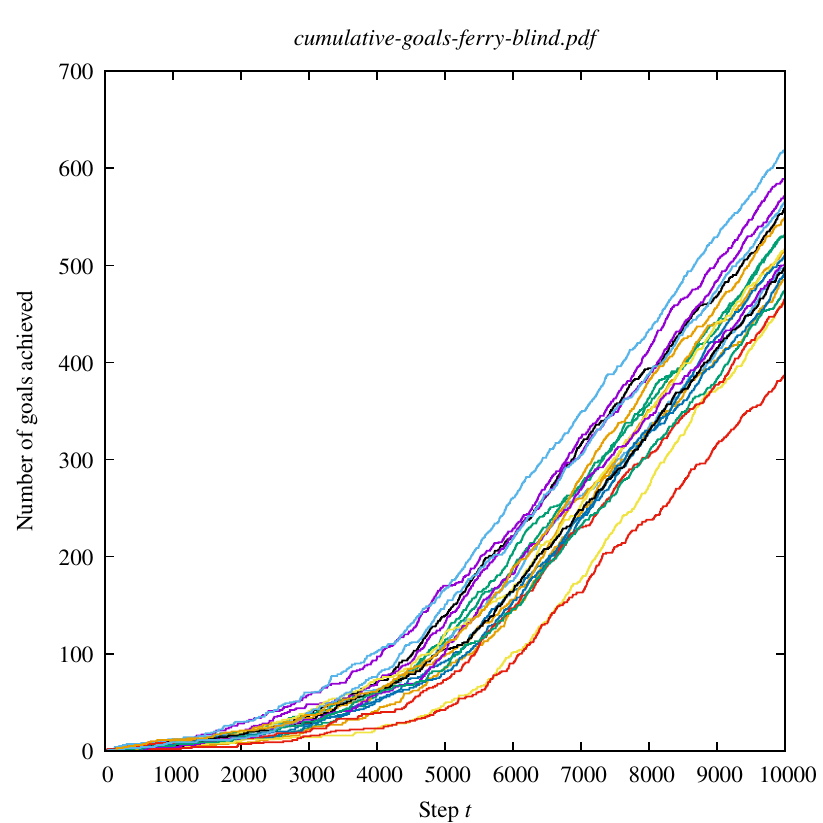}
\includegraphics[width=0.33\linewidth]{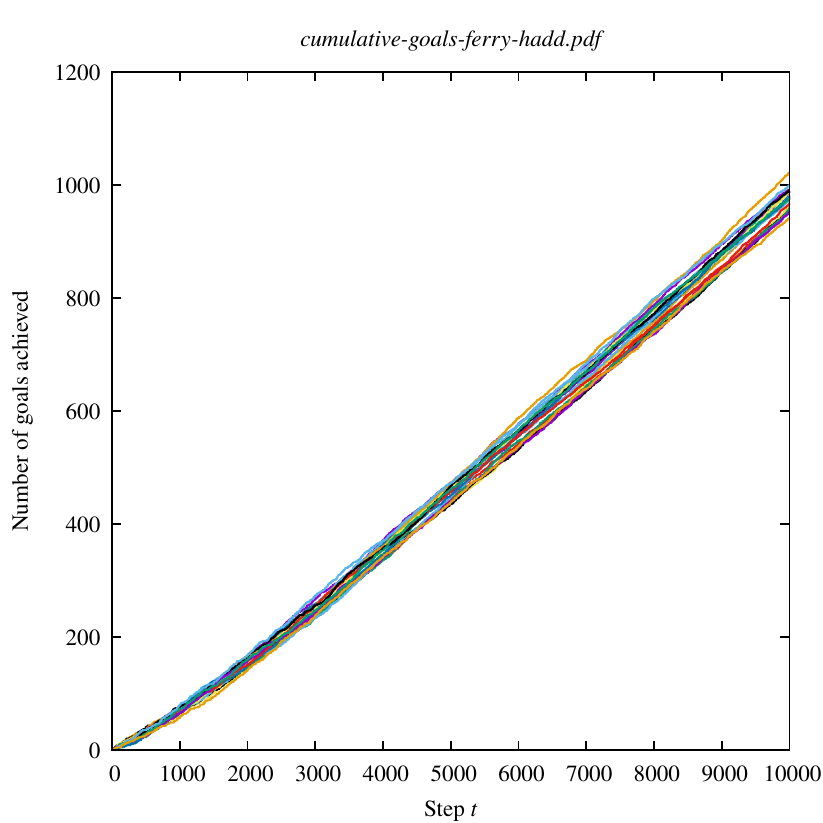}
\includegraphics[width=0.33\linewidth]{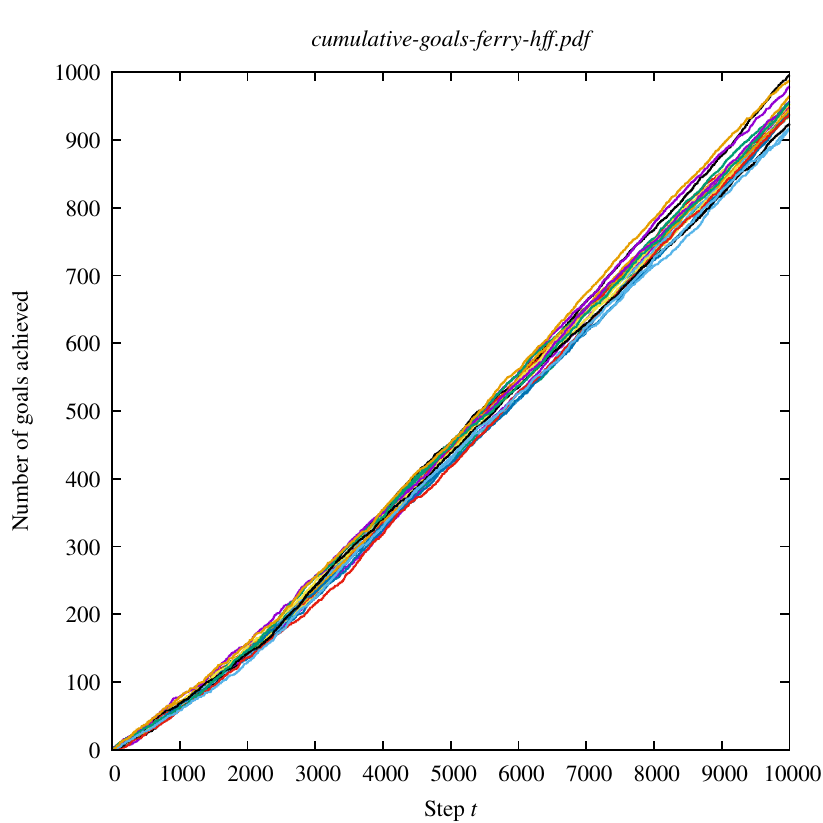}
\includegraphics[width=0.33\linewidth]{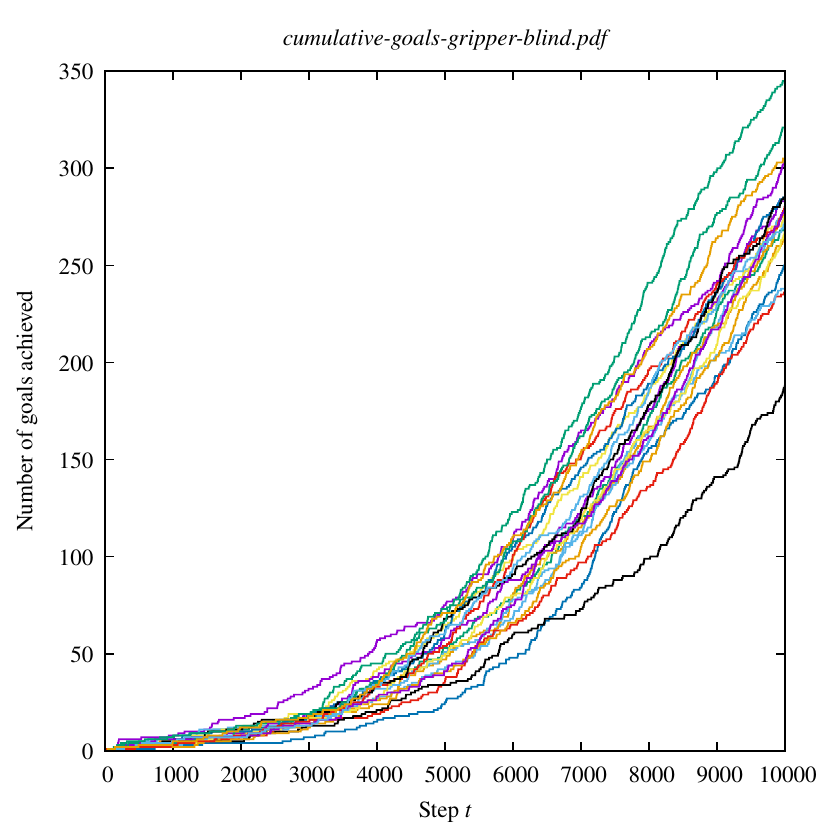}
\includegraphics[width=0.33\linewidth]{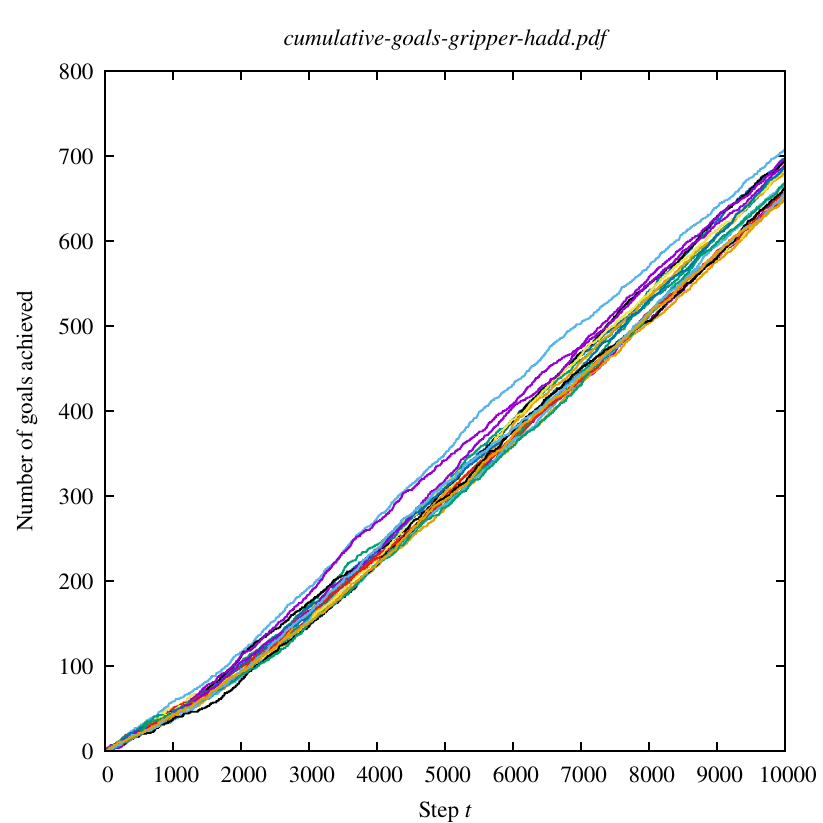}
\includegraphics[width=0.33\linewidth]{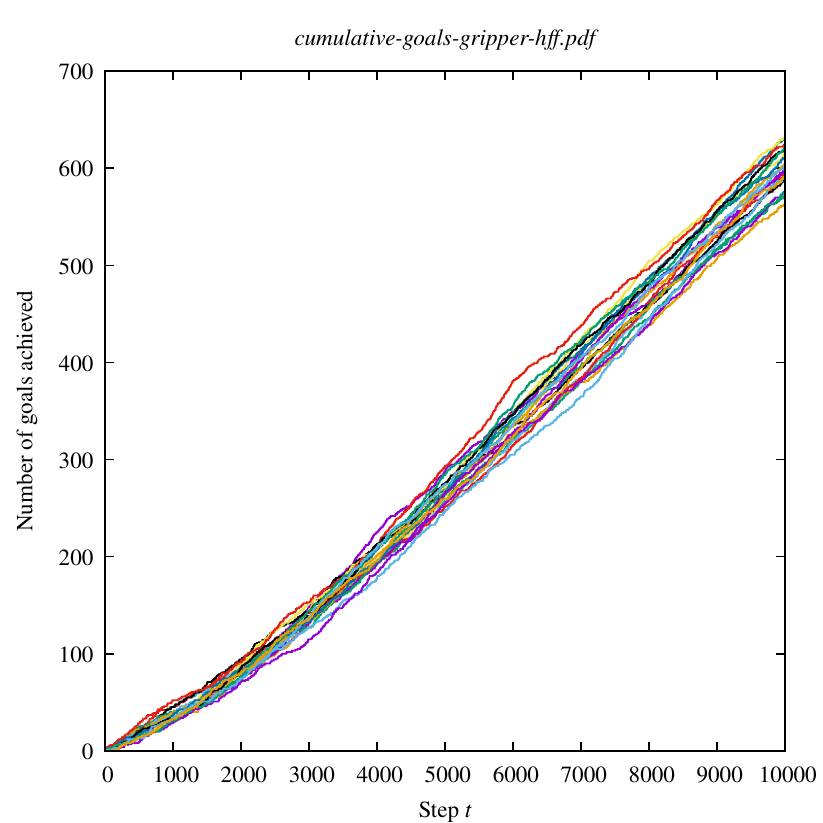}
\includegraphics[width=0.33\linewidth]{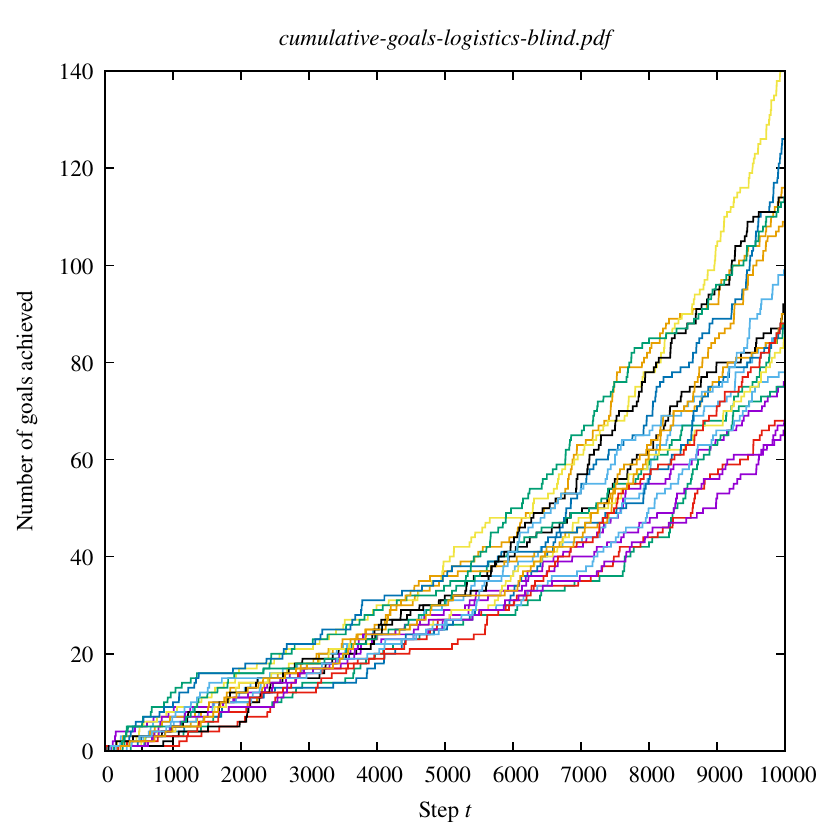}
\includegraphics[width=0.33\linewidth]{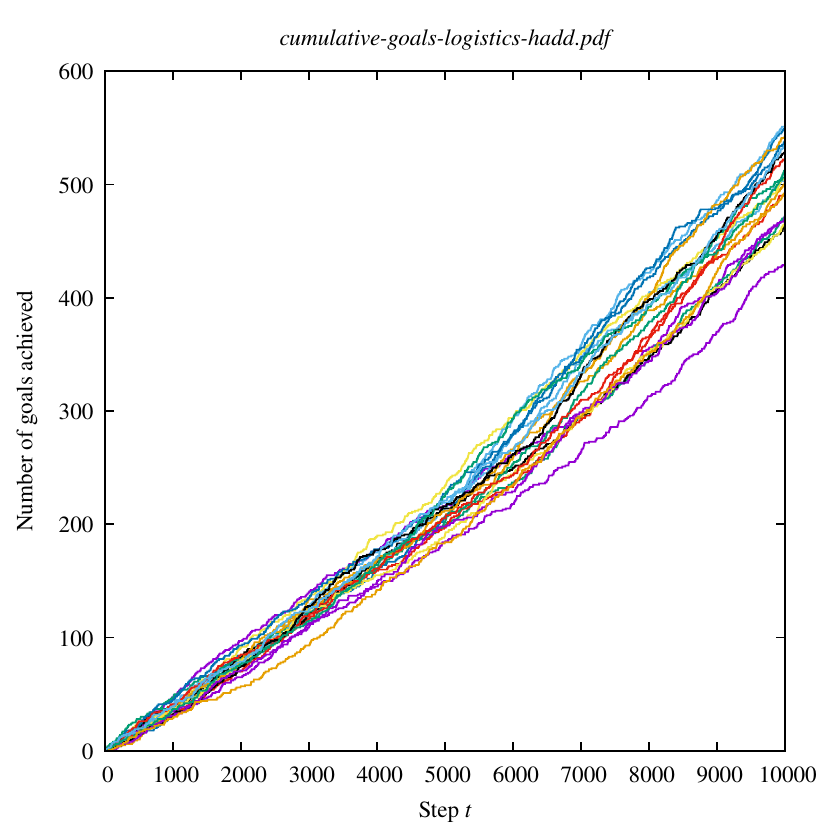}
\includegraphics[width=0.33\linewidth]{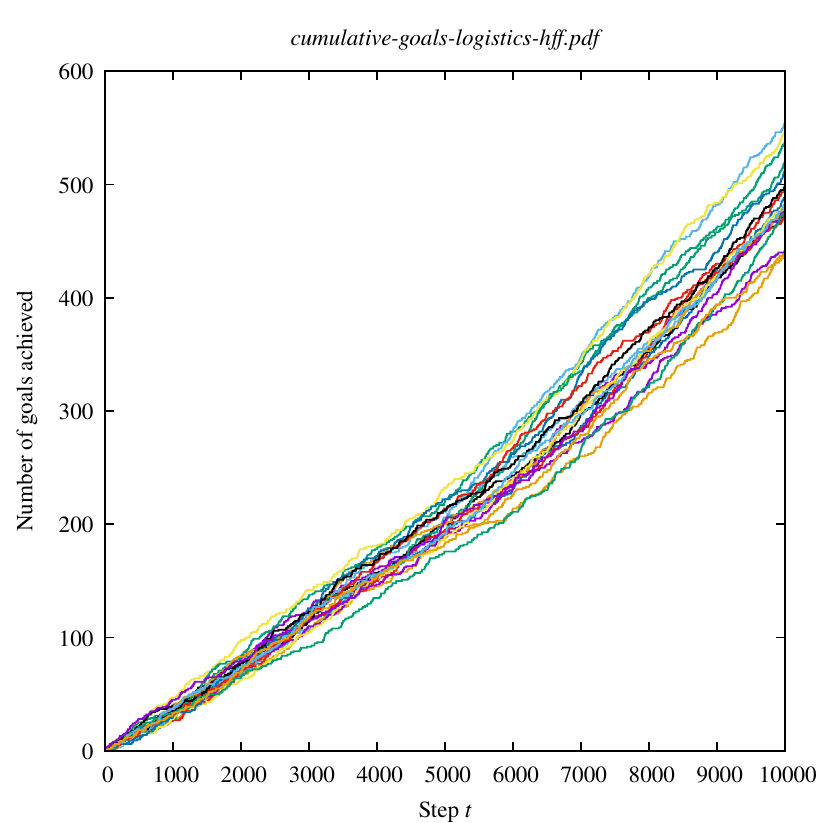}
\caption{
Cumulative number of instances that are solved during the training, where $x$-axis is the training step (part 1).
Note that this may include solving the same instance multiple times.
}
\label{fig:goals1}
\end{figure*}
\begin{figure*}[tb]
\includegraphics[width=0.33\linewidth]{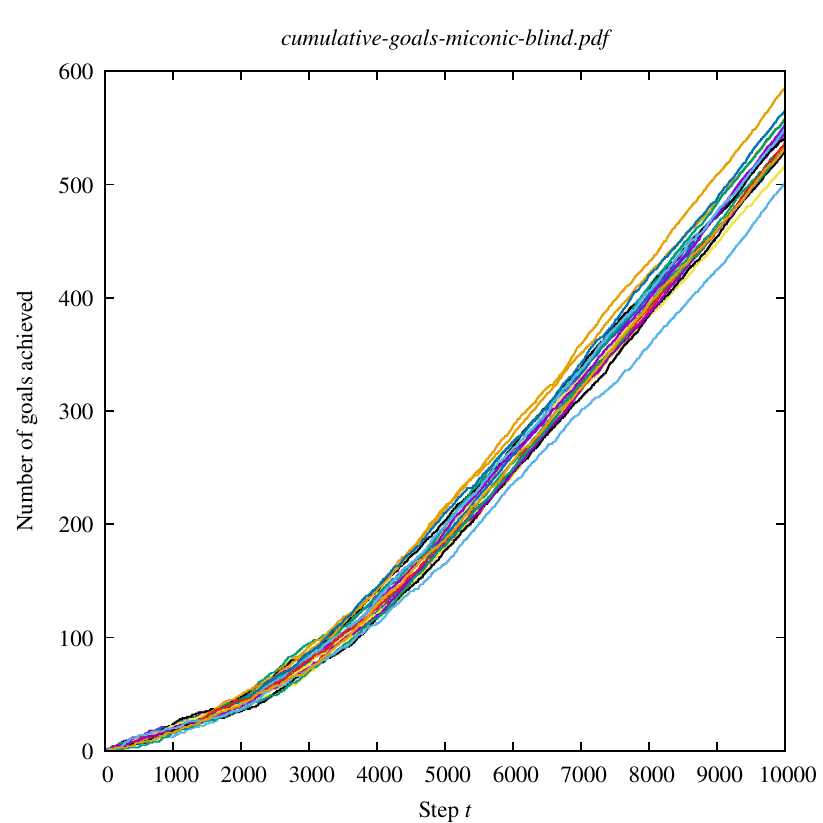}
\includegraphics[width=0.33\linewidth]{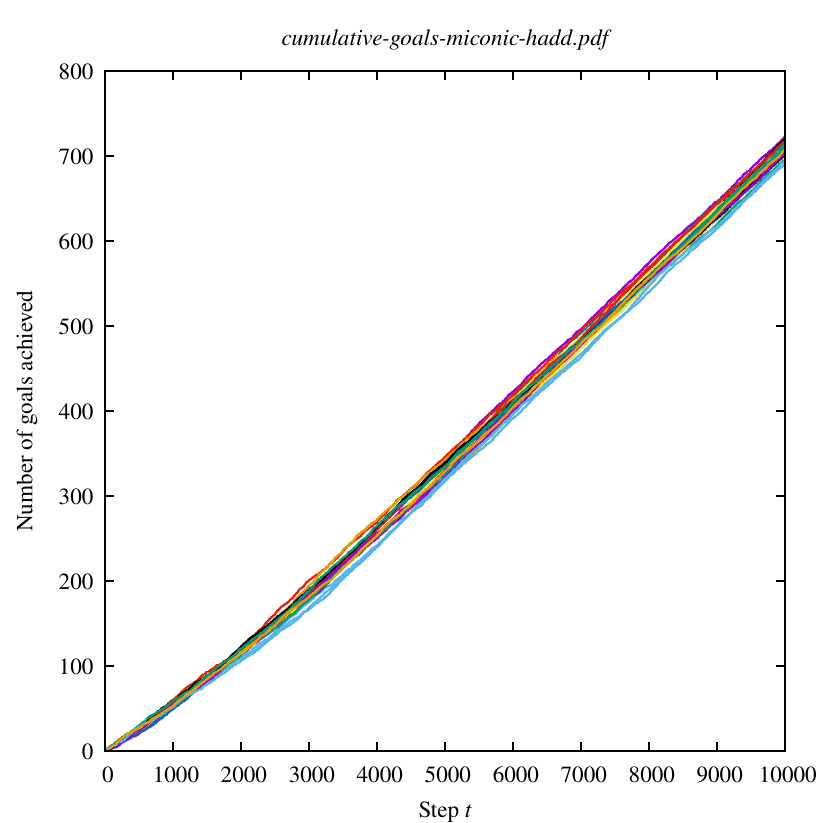}
\includegraphics[width=0.33\linewidth]{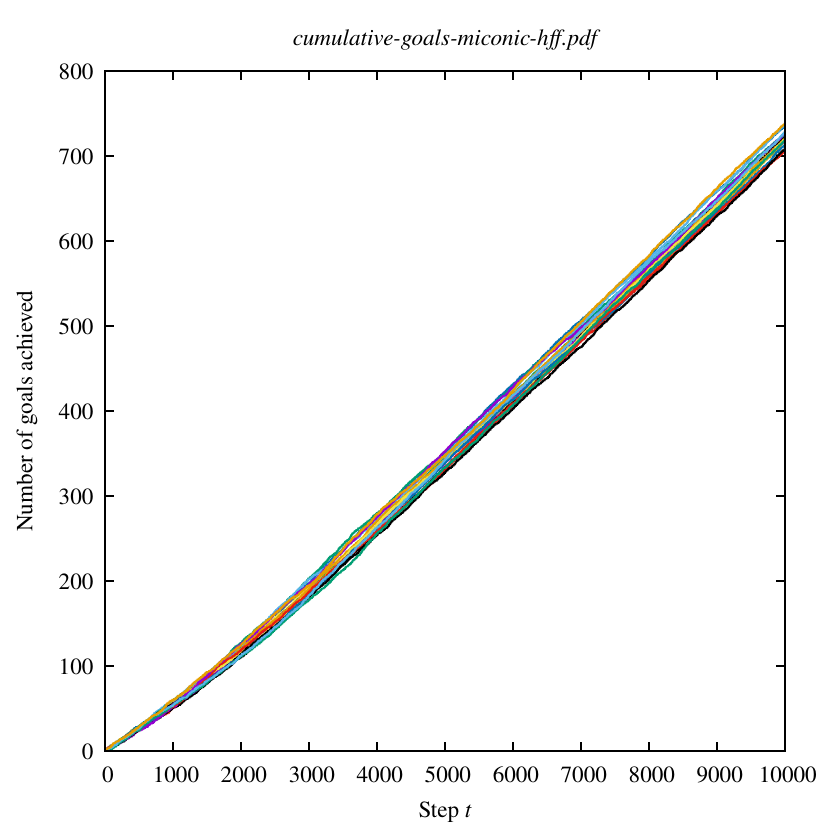}
\includegraphics[width=0.33\linewidth]{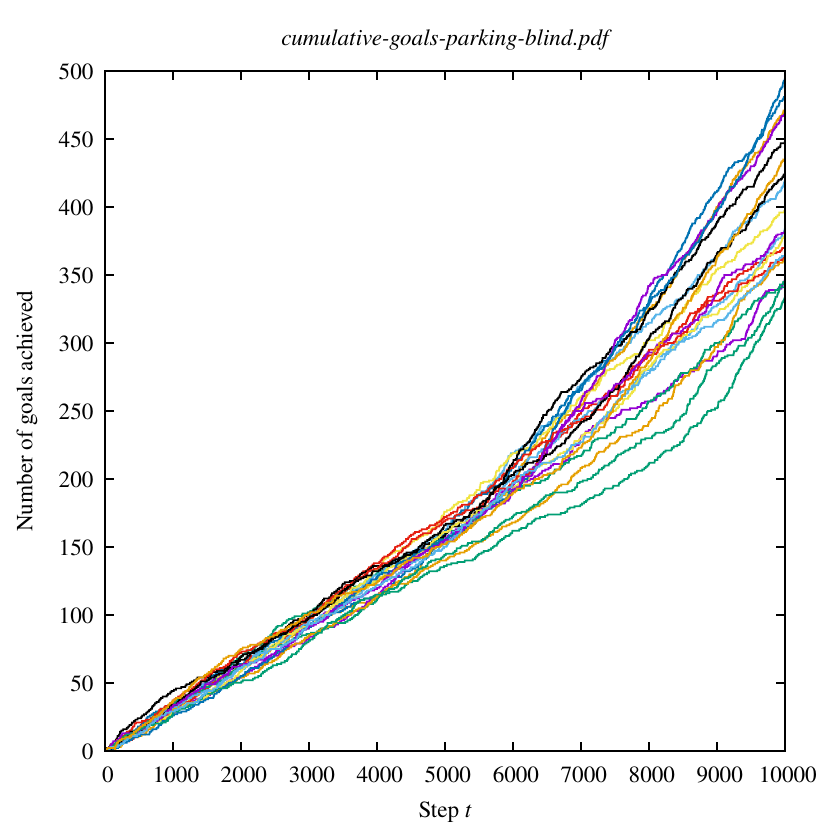}
\includegraphics[width=0.33\linewidth]{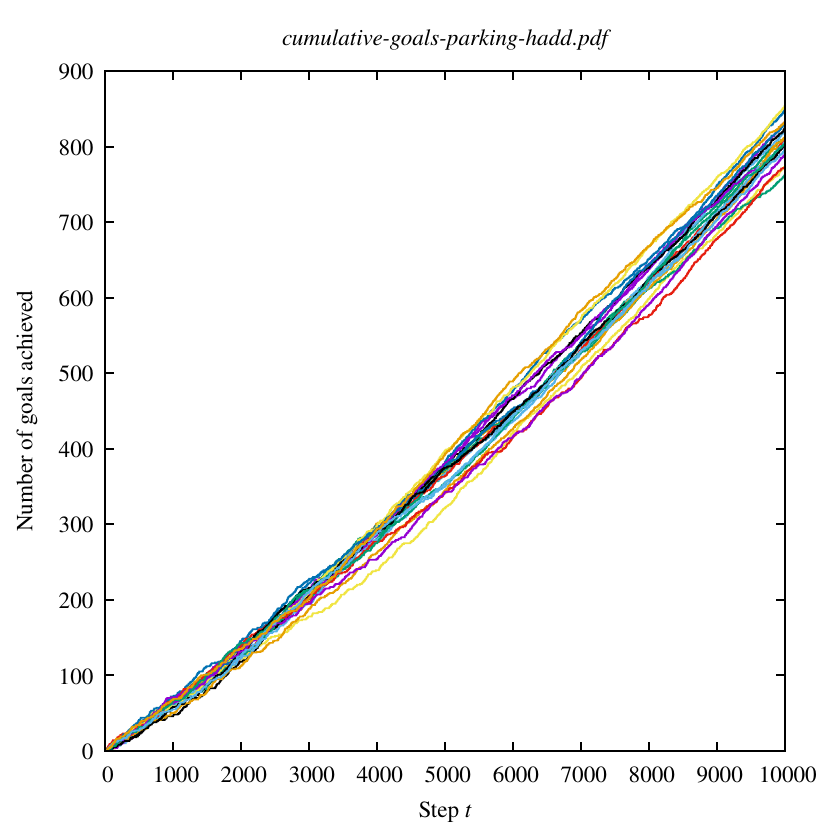}
\includegraphics[width=0.33\linewidth]{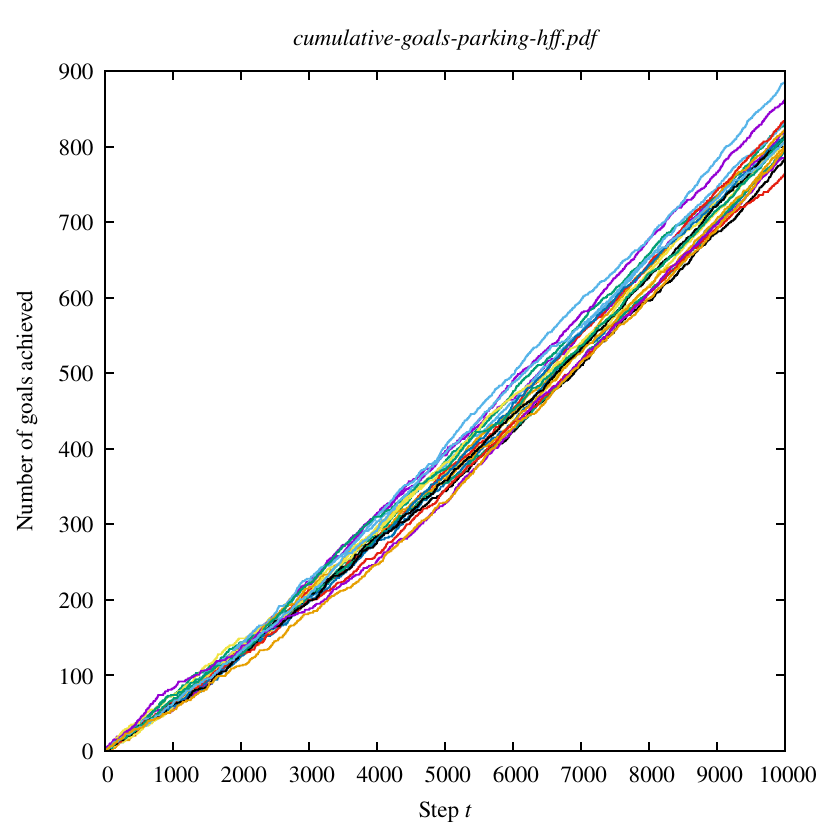}
\includegraphics[width=0.33\linewidth]{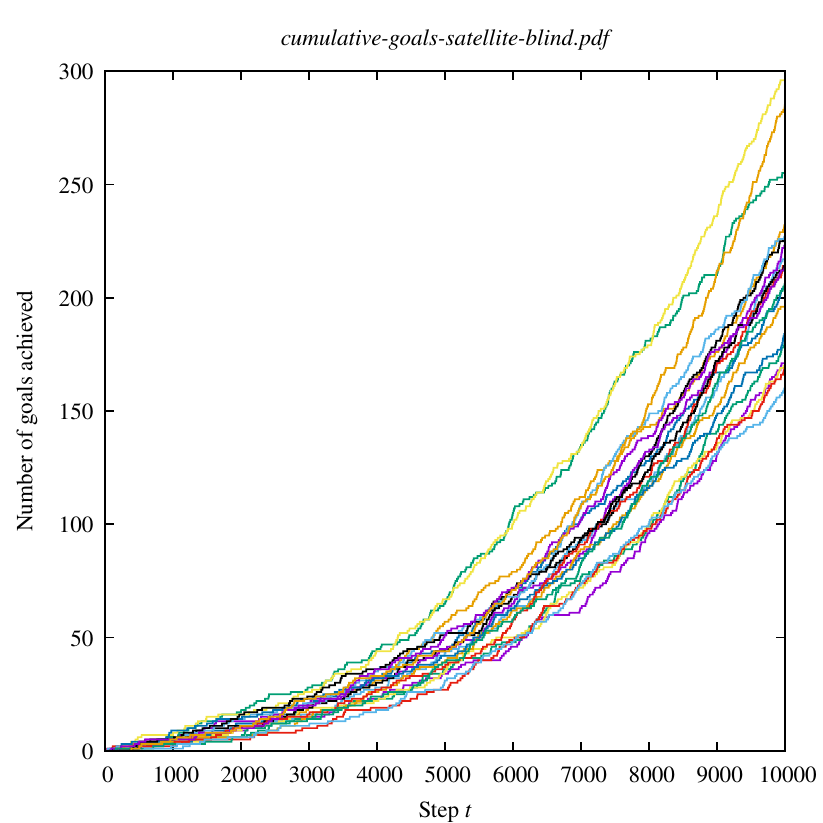}
\includegraphics[width=0.33\linewidth]{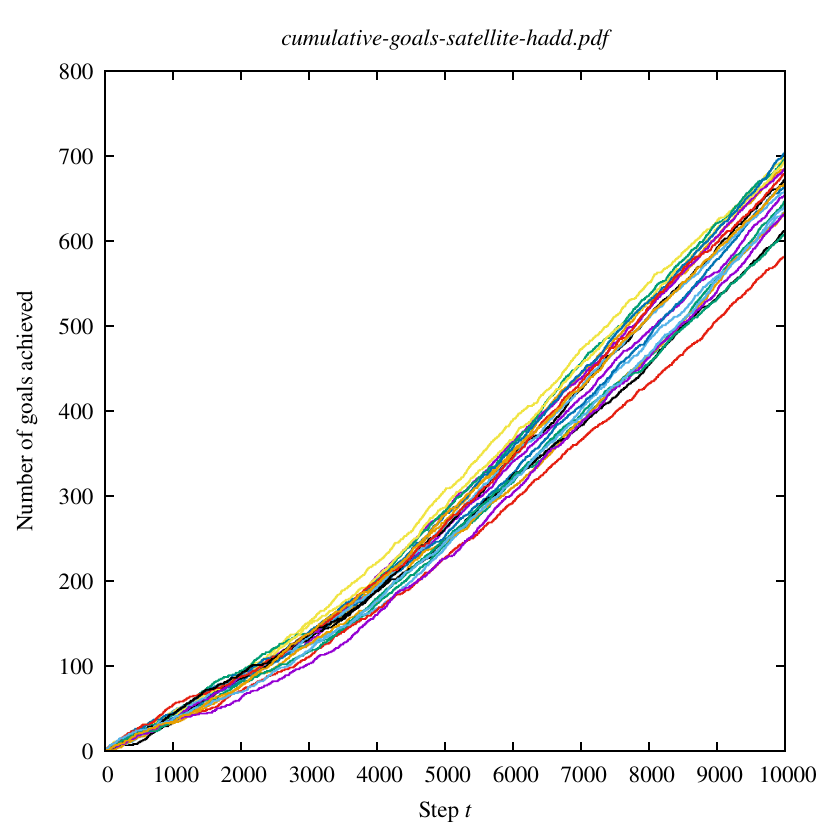}
\includegraphics[width=0.33\linewidth]{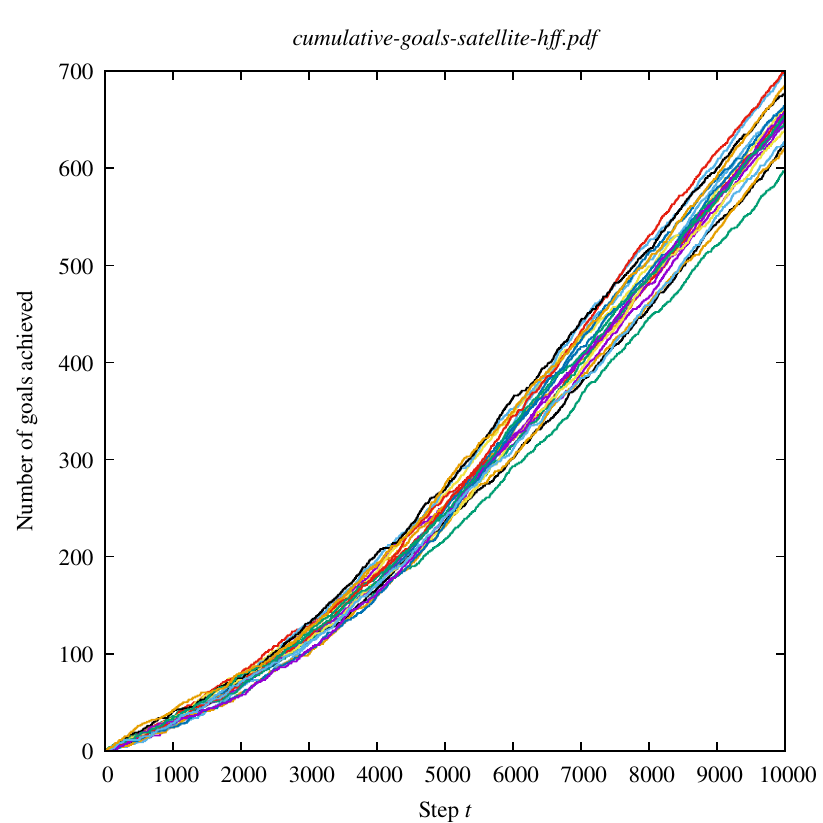}
\includegraphics[width=0.33\linewidth]{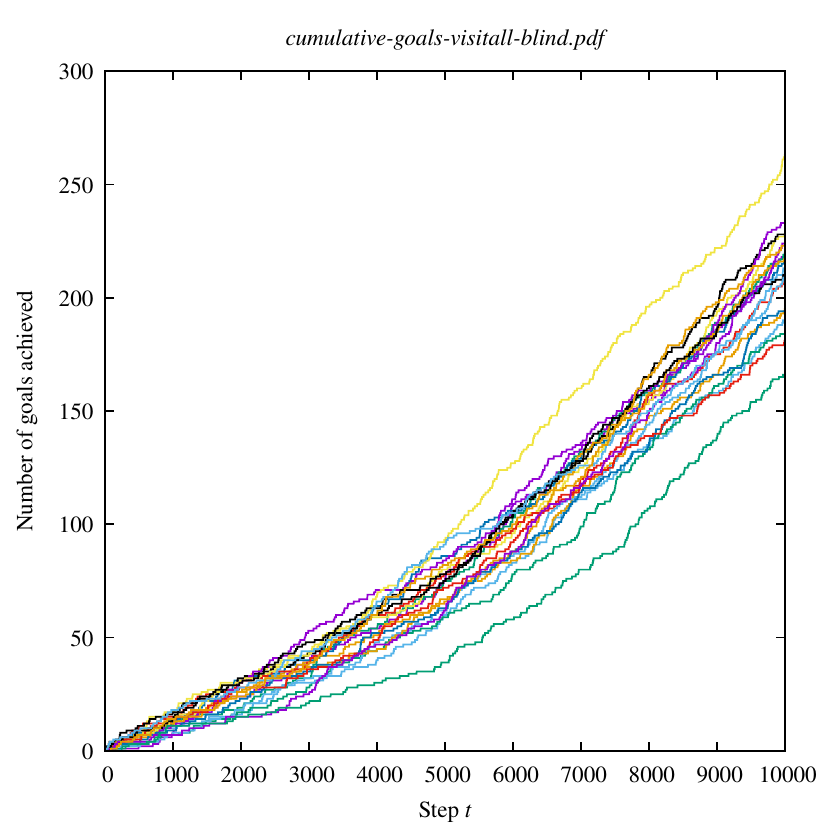}
\includegraphics[width=0.33\linewidth]{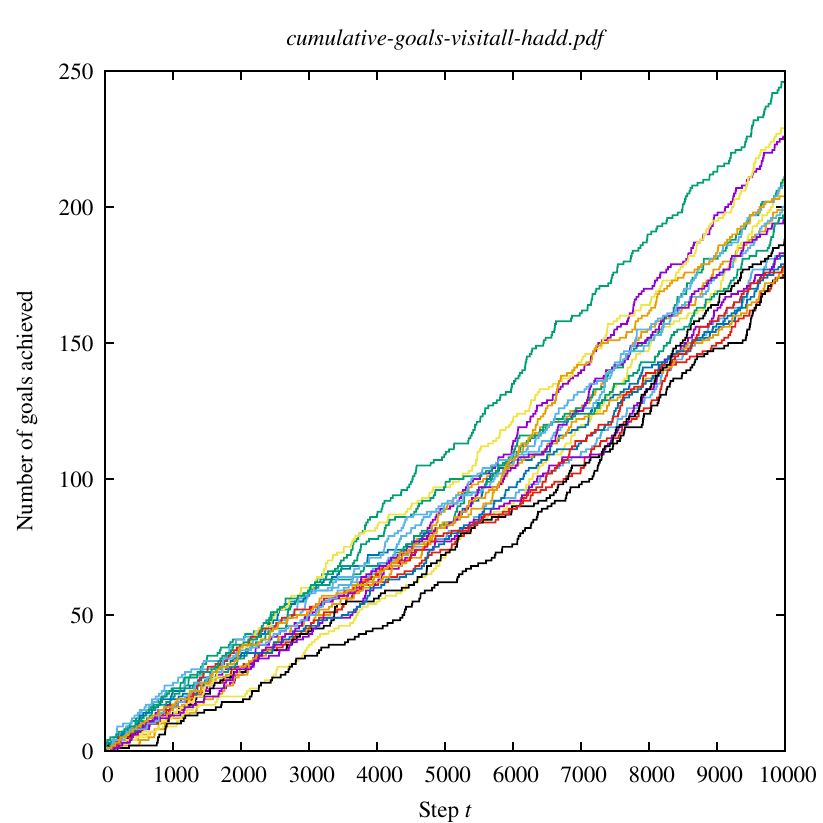}
\includegraphics[width=0.33\linewidth]{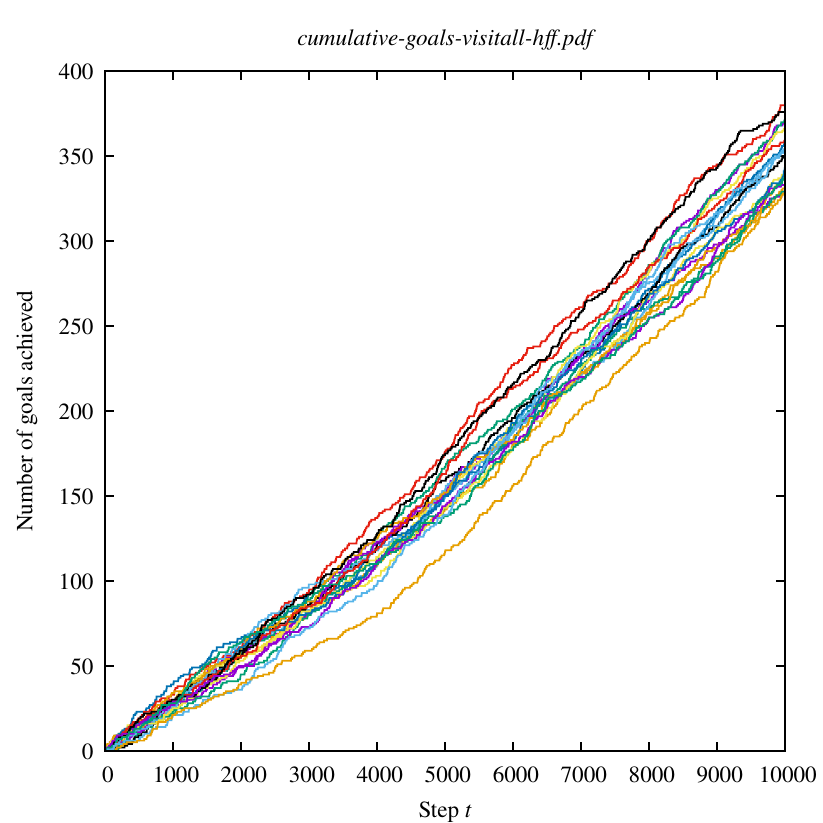}
\caption{
Cumulative number of instances that are solved during the training, where $x$-axis is the training step (part 2).
Note that this may include solving the same instance multiple times.
}
\label{fig:goals2}
\end{figure*}

\end{document}